\documentclass[nohyperref]{article}

\usepackage{microtype}
\usepackage{graphicx}
\usepackage{booktabs} 
\usepackage{subcaption}
\usepackage{hyperref}

\usepackage[accepted]{icml2023}

\usepackage{amsmath}
\usepackage{amssymb}
\usepackage{mathtools}
\usepackage{amsthm}

\usepackage[capitalize,noabbrev]{cleveref}

\theoremstyle{plain}

\theoremstyle{definition}

\theoremstyle{remark}

\usepackage[textsize=tiny]{todonotes}

\usepackage{graphicx}
\usepackage{amsmath}
\usepackage{amssymb}
\usepackage{booktabs}

\usepackage{amsmath}
\usepackage{diagbox}
\usepackage{graphics}
\usepackage{epsfig}
\usepackage{threeparttable}
\usepackage{xspace}
\usepackage{siunitx}
\usepackage{graphicx}
\usepackage{amssymb}
\usepackage{threeparttable}
\usepackage{wrapfig}
\usepackage{color}
\usepackage[normalem]{ulem}
\usepackage{multirow}
\usepackage{float}
\usepackage{amsfonts}
\usepackage{bm}
\usepackage{array}
\usepackage{colortbl}
\usepackage{diagbox}
\usepackage{rotating}
\usepackage{booktabs}
\usepackage{overpic}
\usepackage{enumitem}
\usepackage{colortbl}
\usepackage[export]{adjustbox}
\usepackage{listings}
\usepackage{pifont}
\usepackage{makecell}
\usepackage{arydshln}
\definecolor{mygray}{gray}{.92}
\renewcommand{\algorithmicrequire}{\textbf{Input:}}
\renewcommand{\algorithmicensure}{\textbf{Output:}}

\DeclareMathOperator*{\argmax}{argmax}

\definecolor{mygray}{gray}{.9}
\definecolor{ggray}{RGB}{127,127,127}
\definecolor{reda}{RGB}{192,0,0}
\definecolor{redb}{RGB}{217,148,143}
\definecolor{myyellow}{RGB}{190,144,0}
\definecolor{mygreen}{RGB}{80,100,40}
\definecolor{myblue}{RGB}{30,90,100}
\definecolor{codegreen}{RGB}{79,126,127}
\definecolor{codedefine}{RGB}{153,54,159}
\definecolor{codefunc}{RGB}{73,122,234}
\definecolor{codecall}{RGB}{73,122,234}
\definecolor{codepro}{RGB}{212,96,80}
\definecolor{codedim}{RGB}{89,152,195}
\renewcommand{\algorithmicrequire}{ \textbf{Input:}} 
\renewcommand{\algorithmicensure}{ \textbf{Output:}} 

\definecolor{mygreen2}{RGB}{80,100,40}   

\usepackage{amsthm}
\usepackage{tabulary}
\newcolumntype{I}{!{\vrule width 1pt}}
\newcolumntype{x}[1]{>{\centering\arraybackslash}p{#1pt}}
\newcolumntype{y}[1]{>{\raggedright\arraybackslash}p{#1pt}}
\newcolumntype{z}[1]{>{\raggedleft\arraybackslash}p{#1pt}}

\definecolor{mygreen3}{HTML}{39b54a}   

\definecolor{mydarkblue}{rgb}{0,0.08,0.45}
\hypersetup{hidelinks,colorlinks=true,
linkcolor=mydarkblue,
citecolor=mydarkblue}

\makeatletter
\newcommand{\thickhline}{%
	\noalign {\ifnum 0=`}\fi \hrule height 1pt
	\futurelet \reserved@a \@xhline
}
\makeatother

\makeatletter
\DeclareRobustCommand\onedot{\futurelet\@let@token\@onedot}
\def\@onedot{\ifx\@let@token.\else.\null\fi\xspace}

\def\eg{\emph{e.g}\onedot} 
\def\ie{\emph{i.e}\onedot}

\def\wrt{w.r.t\onedot}

\makeatother

\newcommand{\cmark}{\ding{51}}%
\usepackage[capitalize]{cleveref}
\crefname{section}{Sec.}{Secs.}
\Crefname{section}{Section}{Sections}
\Crefname{table}{Table}{Tables}
\crefname{table}{Tab.}{Tabs.}

\icmltitlerunning{\textsc{ClustSeg}: Clustering for Universal Segmentation}

\begin{document}

\twocolumn[
\icmltitle{\textsc{ClustSeg}: Clustering for Universal Segmentation}




\begin{icmlauthorlist}
\icmlauthor{James Liang}{rit}
\icmlauthor{Tianfei Zhou}{eth}
\icmlauthor{Dongfang Liu}{rit}
\icmlauthor{Wenguan Wang}{zu}
\end{icmlauthorlist}
\center{\url{https://github.com/JamesLiang819/ClustSeg}}
\icmlaffiliation{rit}{Rochester Institute of Technology}
\icmlaffiliation{eth}{ETH Zurich}
\icmlaffiliation{zu}{Zhejiang University}

\icmlcorrespondingauthor{Wenguan Wang}{wenguanwang.ai@gmail.com}
\icmlcorrespondingauthor{Dongfang Liu}{dongfang.liu@rit.edu}

\icmlkeywords{Machine Learning, ICML}

\vskip 0.3in
]



\printAffiliationsAndNotice{}

\begin{abstract}
We$_{\!}$ present$_{\!}$ {\textsc{ClustSeg}},$_{\!}$ a$_{\!}$ general,$_{\!}$ transformer-based framework that tackles different image~seg- mentation tasks (\ie, superpixel, semantic, ins- tance, and panoptic) through a unified, neural~clus- tering$_{\!}$ scheme.$_{\!}$ Regarding$_{\!}$ queries$_{\!}$ as$_{\!}$ cluster$_{\!}$ centers, \textsc{ClustSeg}$_{\!}$ is$_{\!}$ innovative$_{\!}$ in$_{\!}$ two$_{\!}$ aspects:$_{\!_{\!}}$ {\ding{172}}$_{\!_{\!}}$ cluster$_{\!}$ centers$_{\!}$ are$_{\!}$ initialized$_{\!}$ in$_{\!}$ heterogeneous$_{\!}$ ways$_{\!}$ so$_{\!}$ as to$_{\!}$ pointedly$_{\!}$ address$_{\!}$ task-specific$_{\!}$ demands$_{\!}$ (\eg, instance- or category-level distinctiveness), yet without modifying the architecture; and {\ding{173}} pixel-cluster assignment, formalized in a cross-attention fashion,$_{\!}$ is$_{\!}$ alternated$_{\!}$ with$_{\!}$ cluster$_{\!}$ center$_{\!}$ update,$_{\!}$~yet without$_{\!}$ learning$_{\!}$ additional$_{\!}$ parameters.$_{\!}$ These$_{\!}$ in- novations$_{\!}$ closely$_{\!}$ link$_{\!}$ {\textsc{ClustSeg}}$_{\!}$ to$_{\!}$
EM$_{\!}$ cluster- ing$_{\!}$ and$_{\!}$ make$_{\!}$ it$_{\!}$ a$_{\!}$ transparent$_{\!}$ and$_{\!}$ powerful$_{\!}$ frame- work$_{\!}$ that$_{\!}$ yields$_{\!}$ superior$_{\!}$ results$_{\!}$ across$_{\!}$ the$_{\!}$ above segmentation tasks. 

\end{abstract}
	\vspace{-10pt}
\section{Introduction}
\label{sec:introduction}
Image segmentation aims at partitioning pixels into groups. Different notions of pixel groups lead to different types of segmentation tasks. For example, \textit{superpixel} segmentation groups perceptually similar and spatially coherent pixels to- gether.$_{\!}$ \textit{Semantic}$_{\!}$ and$_{\!}$ \textit{instance}$_{\!}$ segmentation$_{\!}$ interpret$_{\!}$ pixel$_{\!}$ groups based on semantic and instance relations respectively. \textit{Panoptic} segmentation$_{\!}$~\cite{kirillov2019panoptic}~not~only distinguishes pixels for countable \textit{things} (\eg, dog, car) at
the instance level, but merges pixels of amorphous and uncountable \textit{stuff} regions (\eg, sky, grassland) at the semantic level.

These segmentation tasks are traditionally resolved by different technical protocols, \eg, \textit{per-pixel classification} for semantic segmentation, \textit{detect-then-segment} for$_{\!}$ instance$_{\!}$ segmentation,$_{\!}$ and$_{\!}$ \textit{proxy$_{\!}$ task$_{\!}$ learning}$_{\!}$ for$_{\!}$ panoptic segmentation. As a result, the developed segmentation solutions are highly {task-specialized}, and research endeavors are diffused.

\begin{figure}[t]
	\begin{center}
		\includegraphics[width=\linewidth]{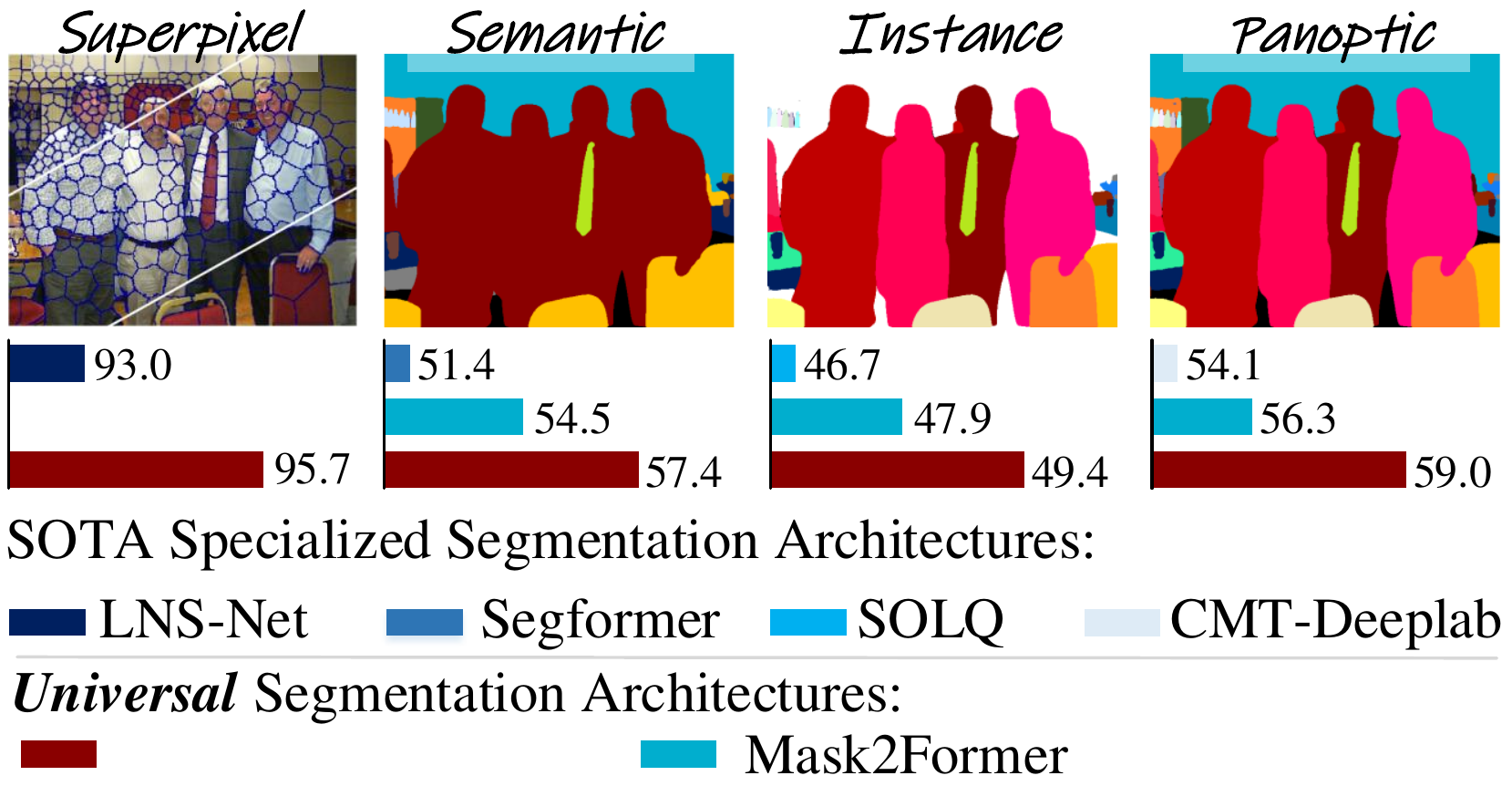}
    \put(-216,1.5){\footnotesize \textsc{ClustSeg} (\textbf{ours}) }
	\end{center}
	\vspace{-10pt}
	\captionsetup{font=small}
	\caption{\small \textsc{ClustSeg} unifies four segmentation tasks (\ie, superpixel, semantic, instance, and panoptic) from the clustering view, and greatly suppresses existing specialized and unified models.}
	\vspace{-12pt}
	\label{fig:1}
\end{figure}

To advance the segmentation field in synergy, a paradigm shift from \textit{task-specialized network architectures} towards~a \textit{universal$_{\!}$ framework}$_{\!}$ is$_{\!}$ needed.$_{\!}$ In$_{\!}$ an$_{\!}$ effort$_{\!}$ to$_{\!}$ embrace$_{\!}$ this shift, we propose \textsc{ClustSeg} which unifies four segmentation tasks \textit{viz}. superpixel, semantic, instance, and panoptic segmentation, from$_{\!}$ the$_{\!}$ \textit{clustering}$_{\!}$ perspective$_{\!}$ using$_{\!}$ trans- formers.$_{\!}$ The$_{\!}$ idea of \textit{segment-by-clustering} --- clustering pixels with similar attributes together to form segmentation masks$_{\!}$ ---$_{\!}$ has$_{\!}$ a$_{\!}$ long$_{\!}$ history$_{\!}$~\cite{coleman1979image},$_{\!}$~yet gets$_{\!}$ largely$_{\!}$ overlooked$_{\!}$ nowadays.$_{\!}$ By$_{\!}$ revisiting$_{\!}$ this$_{\!}$ classic idea$_{\!}$ and$_{\!}$~recasting$_{\!}$ the$_{\!}$ cross-attention$_{\!}$ function$_{\!}$ as$_{\!}$ an$_{\!}$ EM$_{\!}$  clus- tering$_{\!}$ calculator,$_{\!}$ \textsc{ClustSeg}$_{\!}$ sticks$_{\!}$ the$_{\!}$ principle$_{\!}$ of$_{\!}$ pixel$_{\!}$ clus- tering through several innovative algorithmic designs, out- performing$_{\!}$ existing$_{\!}$ specialized$_{\!}$ and$_{\!}$ unified$_{\!}$ models$_{\!}$  (Fig.$_{\!}$~\ref{fig:1}).

Concretely, our innovations are centred around two aspects and respect some critical rules of iterative/EM clustering:\\  {\ding{172}}$_{\!}$ \textbf{Cluster$_{\!}$ center$_{\!}$ initialization}:$_{\!\!}$ By resorting to the cross-attention for pixel-cluster assignment, the queries in transformers are  deemed  as cluster centers. From the clustering standpoint, 
the choice of initial centers is of great importance. However, existing transformer-based segmenters sim- ply$_{\!}$ learn$_{\!}$ the$_{\!}$ queries$_{\!}$ in$_{\!}$ a$_{\!}$ fully$_{\!}$ parametric$_{\!}$ manner.$_{\!}$ By$_{\!}$ res- pecting$_{\!}$ task-specific$_{\!}$ natures,
\textsc{ClustSeg} implants concrete meanings to
queries:$_{\!}$ for semantic/stuff segmentation, they are invented as class centers (as the semantic membership is defined on the category level), whereas queries for superpixels/instances/things are emerged purely from the individual input image (as the target tasks are scene-/instance-specific). This$_{\!}$ smart$_{\!}$ {query-initialization$_{\!}$ scheme},$_{\!}$ called$_{\!}$ \textit{dreamy-start}, boosts pixel grouping with more informative seeds, as well as allows \textsc{ClustSeg} to accommodate the heterogeneous properties of different tasks into one single architecture.\\
{\ding{173}}$_{\!}$ \textbf{Iterative$_{\!}$ clustering$_{\!}$ and$_{\!}$ center$_{\!}$ update}:$_{\!}$ To$_{\!}$ approximate the$_{\!}$ optimal$_{\!}$ clustering,$_{\!}$ EM$_{\!}$ iteratively$_{\!}$ alters$_{\!}$ cluster$_{\!}$ member-
ship$_{\!}$ and$_{\!}$ centers.$_{\!}$ But$_{\!}$ current$_{\!}$ transformer-based$_{\!}$ segmenters only$_{\!}$
update$_{\!}$ the$_{\!}$ query$_{\!}$ centers$_{\!}$ via$_{\!}$ a$_{\!}$ few$_{\!}$ cross-attention$_{\!}$ based$_{\!}$
decoders$_{\!}$  (typically$_{\!}$  six$_{\!}$ \cite{cheng2021per}).$_{\!}$ Given$_{\!}$ the$_{\!}$ suc-
cess$_{\!}$ of EM$_{\!}$ clustering,$_{\!}$ we$_{\!}$ devise$_{\!}$ \textit{recurrent$_{\!}$ cross-attention} that$_{\!}$ repeatedly$_{\!}$ alters$_{\!}$ cross-attention$_{\!}$ computation$_{\!}$ (for$_{\!}$ pixel- cluster$_{\!}$ assignment)$_{\!}$ and$_{\!}$ attention-based$_{\!}$ feature$_{\!}$ aggregation$_{\!}$ (for$_{\!}$ center$_{\!}$ update).$_{\!}$ By$_{\!}$ embedding$_{\!}$ such$_{\!}$ nonparametric$_{\!}$ re- current$_{\!}$ mechanism,$_{\!}$ \textsc{ClustSeg}$_{\!}$ fully$_{\!}$ explores$_{\!}$ the$_{\!}$ power$_{\!}$ of iterative$_{\!}$ clustering$_{\!}$ in$_{\!}$ pixel$_{\!}$ grouping,$_{\!}$ without$_{\!}$ additional$_{\!}$ learnable parameters and discernible  inference speed reduction.

Taking these innovations together, \textsc{ClustSeg} becomes a general, flexible, and transparent framework for image segmentation. Unlike prior mask-classification based universal segmenters~\cite{zhang2021k,cheng2021per,cheng2021masked}, our \textsc{ClustSeg} acknowledges the fundamental principle of segment-by-clustering.$_{\!}$ There$_{\!}$ are$_{\!}$ a$_{\!}$ few$_{\!}$~clustering$_{\!}$ based$_{\!}$~seg- mentation$_{\!}$ networks$_{\!}$~\cite{kong2018recurrent,neven2019instance,yu2022cmt,yu2022k}$_{\!}$ ---$_{\!}$ their successes, though limited in their specific targeting tasks, shed light on the potential of unifying image segmentation as pixel clustering.~\textsc{ClustSeg}, for the first time, shows impressive performance on four core segmentation tasks. In particular, \textsc{ClustSeg} sets tantalizing records of \textbf{59.0} PQ on COCO panoptic segmen- tation$_{\!}$~\cite{kirillov2019panoptic},$_{\!}$ \textbf{49.1}$_{\!}$ AP$_{\!}$~on$_{\!}$ COCO$_{\!}$ instance$_{\!}$ segmentation$_{\!}$~\cite{lin2014microsoft},$_{\!}$ and$_{\!}$ \textbf{57.4}$_{\!}$ mIoU$_{\!}$ on$_{\!}$ ADE20K semantic$_{\!}$ segmentation$_{\!}$~\cite{zhou2017scene},$_{\!}$ and$_{\!}$ reports$_{\!}$ the best ASA and CO curves on BSDS500 superpixel segmen- tation \cite{arbelaez2011contour}.

	\vspace{-5pt}
\section{Related Work}\label{sec:RW}
	\vspace{-3pt}

\noindent\textbf{Semantic Segmentation} interprets high-level semantic concepts of visual stimuli by grouping pixels into different semantic units. Since the proposal of fully convolutional~net- works (FCNs) \cite{long2015fully}, continuous endeavors have been devoted to the design of more powerful FCN-like models, by \eg, aggregating context~\cite{ronneberger2015u,zheng2015conditional,yu2015multi}, incorporating$_{\!}$  neural$_{\!}$ attention \cite{harley2017segmentation,wang2018non,zhao2018psanet,hu2018squeeze,fu2019dual}, conducting contrastive learning$_{\!}$~\cite{wang2021exploring}, revisiting prototype theory$_{\!}$~\cite{zheng2021rethinking,wang2023visual}, and adopting generative models$_{\!}$~\cite{liang2022gmmseg}. Recently, engagement$_{\!}$ with$_{\!}$ advanced$_{\!}$ transformer$_{\!}$~\cite{vaswani2017attention}$_{\!}$ architecture attained wide research attention~\cite{xie2021segformer,strudel2021segmenter,zheng2021rethinking,zhu2021unified,cheng2021per,cheng2021masked,gu2022multi}.

\noindent\textbf{Instance Segmentation} groups foreground pixels into different object instances. There are three types of solutions:~\textbf{i)} \textbf{top-down} models, built upon a  \textit{detect-then-segment} proto- col, first detect object bounding boxes and then delineate an instance mask for each box~\cite{he2017mask,chen2018masklab,huang2019mask,cai2019cascade,chen2019hybrid}; \textbf{ii)} \textbf{bottom-up} models learn instance-specific pixel embeddings by considering, \eg, instance boundaries \cite{kirillov2017instancecut}, energy levels \cite{bai2017deep}, geometric structures \cite{chen2019tensormask}, and pixel-center offsets \cite{zhou2021differentiable}, and then merge them as instances; and \textbf{iii)} \textbf{single-shot} approaches directly predict instance masks by locations using a set of learnable object queries \cite{wang2020solo,wang2020solov2,fang2021instances,guo2021sotr,dong2021solq,hu2021istr,cheng2022sparse,wang2022learning}.

\vspace{-1pt}
\noindent\textbf{Panoptic$_{\!}$ Segmentation}$_{\!}$ seeks$_{\!}$ for$_{\!}$ holistic$_{\!}$ scene$_{\!}$ understanding, in terms of the semantic relation between background stuff pixels and the instance membership between foreground thing pixels. Starting from the pioneering
work~\cite{kirillov2019panoptic}, prevalent solutions \cite{kirillov2019panopticfpn,xiong2019upsnet,li2019attention,liu2019end,lazarow2020learning,li2020unifying,wang2020pixel} decompose the problem into various manageable proxy tasks, including box detection, box-based
segmentation, and thing-stuff merging. Later,  DETR \cite{carion2020end}  and Panoptic FCN~\cite{li2021fully} led a shift towards$_{\!}$ end-to-end$_{\!}$ panoptic$_{\!}$ segmentation$_{\!}$~\cite{cheng2020panoptic,wang2020axial,wang2021max,yu2022cmt,yu2022k}. These
compact panoptic architectures show the promise of unifying semantic and instance segmentation, but are usually sub-optimal compared with specialized models. This calls for endeavor of more powerful universal algorithms for segmentation.

\vspace{-1pt}
\noindent\textbf{Superpixel Segmentation} is to give a concise image repre- sentation by grouping pixels into perceptually meaningful small patches (\ie, superpixel). Superpixel segmentation is an active research area in the pre-deep learning~era; see \cite{stutz2018superpixels} for a thorough survey. Recently, some approaches are developed to harness neural networks~to~fa- cilitate superpixel segmentation~\cite{jampani2018superpixel,yang2020superpixel,zhu2021learning}. For instance,~\Citet{tu2018learning} make use of deep learning techniques to learn a superpixel-friendly embedding space; \Citet{yang2020superpixel} adopt a FCN to directly predict association scores between pixels and regular grid cells for grid-based superpixel creation.

\vspace{-1pt}
\noindent\textbf{Universal$_{\!}$ Image$_{\!}$ Segmentation}$_{\!}$ pursues a unified architecture for tackling different segmentation tasks. Existing~task-specific segmentation models, though advancing the~perfor- mance in their individual tasks, lack flexibility to generalize to other tasks and cause duplicate research effort.~\Citet{zhang2021k} initiate the attempt to unify~segmentation by dynamic$_{\!}$ kernel$_{\!}$ learning.$_{\!}$ More$_{\!}$ recently,$_{\!}$ \Citet{cheng2021per,cheng2021masked} formulate different tasks within a mask-classification scheme, using a transformer decoder with object queries. Compared with these pioneers, \textsc{ClustSeg} is \textbf{i)} more \textit{transparent} and \textit{insightful} --- it explicitly acknowledges the fundamental and easy-to-understand principle of segment-by-clustering;$_{\!}$ \textbf{ii)}$_{\!}$ more$_{\!}$ \textit{versatile}$_{\!}$ ---$_{\!}$ it$_{\!}$ handles$_{\!}$ more$_{\!}$ segmentation tasks$_{\!}$ unanimously;$_{\!}$ \textbf{iii)}$_{\!}$ more$_{\!}$ \textit{flexible}$_{\!}$ ---$_{\!}$ it$_{\!}$ respects,$_{\!}$ instead$_{\!}$ of ignoring, the divergent characters of different segmentation tasks; and \textbf{iv)} more  \textit{powerful} --- it leads by large margins.

\noindent\textbf{Segmentation-by-Clustering},$_{\!}$ a$_{\!}$ once$_{\!}$ popular$_{\!}$ paradigm,$_{\!}$ received far less attention nowadays. Recent investigations~of such paradigm are primarily made around bottom-up ins- tance segmentation$_{\!}$~\cite{kong2018recurrent,neven2019instance}, where the clustering is adopted as a post-processing step after learning an instance-aware pixel embedding space. More recently, \Citet{yu2022cmt,yu2022k} build end-to-end, clustering based panoptic systems by reformulating the cross-attention as a clustering solver. In this work, we further advance this research line for the development of universal image segmentation. With the innovations in task-specific cluster center initialization and nonparametric recurrent cross-attention,$_{\!}$ \textsc{ClustSeg}$_{\!}$ better$_{\!}$ adheres$_{\!}$ to$_{\!}$ the$_{\!}$ nature of clustering and elaborately deals with the heterogeneity across different segmentation tasks using the same architecture.

	\vspace{-3pt}
\section{Methodology}
	\vspace{-1pt}

\subsection{Notation and Preliminary}\label{sec:3.1}
	\vspace{-3pt}

\noindent\textbf{Problem Statement.} Image segmentation seeks to partition an image ${I}\!\in\!\mathbb{R}^{H_{\!}W_{\!}\times{\!}3_{\!}}$ into a set of $K$ meaningful segments:
\vspace{-2pt}
\begin{equation}
\begin{aligned}\label{eq:1}
\text{segment}(I)=\{M_k\!\in\!\{0, 1\}^{HW\!}\}_{k=1}^{K},
\end{aligned}
\vspace{-2pt}
\end{equation}
where$_{\!}$ $M_{k}(i)_{\!}$ denotes$_{\!}$ if$_{\!}$ pixel$_{\!}$ $i_{\!}\!\in_{\!}\!I_{\!}$ is$_{\!}$~($1$)$_{\!}$ or$_{\!}$ not$_{\!}$~($0$)$_{\!}$ a$_{\!}$ member of$_{\!}$~segment$_{\!}$ $k$.$_{\!}$ Different$_{\!}$ segmentation$_{\!}$ tasks$_{\!}$ find$_{\!}$ the$_{\!}$ segments according to, for example, semantics, instance membership,$_{\!}$ or$_{\!}$ low-level$_{\!}$ attributes.$_{\!}$ 
Also,$_{\!}$ the$_{\!}$ number$_{\!}$~of$_{\!}$ segments,$_{\!}$ $K$,$_{\!}$~is different$_{\!}$ for$_{\!}$ different$_{\!}$ tasks:$_{\!\!}$ In$_{\!}$ \textit{superpixel}$_{\!}$ segmentation,$_{\!}$ $K_{\!}$~is$_{\!}$~a pre-determined$_{\!}$ arbitrary$_{\!}$ value,$_{\!}$ \ie,$_{\!}$ compress$_{\!}$ $I_{\!}$ into $K_{\!}$ super-
pixels;$_{\!}$ In$_{\!}$ \textit{semantic}$_{\!}$ segmentation,$_{\!}$ $K_{\!}$ is$_{\!}$ fixed$_{\!}$ as$_{\!}$ the$_{\!}$ length$_{\!}$~of$_{\!}$~a pre-given$_{\!}$ list$_{\!}$ of$_{\!}$ semantic$_{\!}$ tags;$_{\!}$ In \textit{instance}$_{\!}$ and$_{\!}$ \textit{panoptic}$_{\!}$ seg- mentation,$_{\!}$ $K_{\!}$ varies$_{\!}$ across$_{\!}$ images$_{\!}$ and$_{\!}$ needs$_{\!}$ to  be inferred, as the number of object instances presented in an image is unknown.$_{\!}$ Some$_{\!}$ segmentation$_{\!}$ tasks$_{\!}$ require$_{\!}$ explaining$_{\!}$ the$_{\!}$ se- mantics$_{\!}$ of$_{\!}$ segments;$_{\!}$ related$_{\!}$ symbols$_{\!}$ are$_{\!}$ omitted$_{\!}$ for$_{\!}$ clarity.

\noindent\textbf{Unifying$_{\!}$ Segmentation$_{\!}$ as$_{\!}$ Clustering.$_{\!}$} Eq.$_{\!}$~\ref{eq:1}$_{\!}$ reveals$_{\!}$ that,$_{\!}$ though$_{\!}$ differing$_{\!}$ in$_{\!}$ the$_{\!}$~definition$_{\!}$ of$_{\!}$ a$_{\!}$ ``meaningful$_{\!}$ segment'', segmentation$_{\!}$ tasks$_{\!}$ can$_{\!}$ be$_{\!}$ essentially$_{\!}$ viewed$_{\!}$ as$_{\!}$ a$_{\!}$ \textit{pixel$_{\!}$ clus- tering}$_{\!}$ process:$_{\!}$ the$_{\!}$ binary$_{\!}$ segment$_{\!}$ mask,$_{\!}$ \ie, $M_{k}$,$_{\!}$ is$_{\!}$ the$_{\!}$~pixel assignment$_{\!}$ matrix$_{\!}$ \wrt $_{\!}k^{th\!}$ cluster.$_{\!}$ With
this$_{\!}$ insight,$_{\!}$ \textsc{ClustSeg}$_{\!}$ advocates$_{\!}$ unifying$_{\!}$ segmentation$_{\!}$ as$_{\!}$ clustering.$_{\!}$ Note$_{\!}$ that recent$_{\!}$ mask-classification-based$_{\!}$ universal$_{\!}$ segmenters \cite{cheng2021per,cheng2021masked}$_{\!}$ do$_{\!}$ not$_{\!}$ knowledge$_{\!}$ the$_{\!}$ rule$_{\!}$ of$_{\!}$ clu- stering.$_{\!}$ As$_{\!}$ the$_{\!}$ segment$_{\!}$ masks$_{\!}$ are$_{\!}$ the$_{\!}$ outcome$_{\!}$ of$_{\!}$ clustering, the$_{\!}$ viewpoint$_{\!}$ of$_{\!}$ segment-by-clustering$_{\!}$  is$_{\!}$ more$_{\!}$ insightful$_{\!}$ and a$_{\!}$ close$_{\!}$ scrutiny$_{\!}$ of$_{\!}$ classical$_{\!}$ clustering$_{\!}$ algorithms$_{\!}$ is$_{\!}$ needed.

\noindent\textbf{EM$_{\!}$ Clustering.$_{\!}$} As$_{\!}$ a$_{\!}$ general$_{\!}$ family$_{\!}$ of$_{\!}$ iterative$_{\!}$ clustering, EM clustering makes$_{\!}$ $K_{_{\!}}$-way$_{\!}$ clustering$_{\!}$~of$_{\!}$~a$_{\!}$
$D$-dimensional  set$_{\!}$ of$_{\!}$ $L_{\!}$ data$_{\!}$ points$_{\!}$ $\bm{X}_{\!}\!=_{\!}\![\bm{x}_1;\cdots\!;\bm{x}_N]\!\in\!\mathbb{R}^{N\!\times\!D\!}_{\!}$ by$_{\!}$ solving:
\vspace{-2pt}
\begin{equation}
\begin{aligned}\label{eq:kmeans}
\max_{\bm{M}\in\{0,1\}^{K\!\times\!N}}\!\!\!\texttt{Tr}(\bm{M}^\top\bm{C}\bm{X}^{\!\top\!}),~~\!\textit{s.t.}\!~~\bm{1}_K\bm{M}=\!\bm{1}_{N}.
\end{aligned}
\vspace{-2pt}
\end{equation}
Here$_{\!}$ $\bm{C}\!=\![\bm{c}_1;\cdots\!;\bm{c}_K]\!\in\!\mathbb{R}^{K\!\times\!D_{\!}}$ is$_{\!}$ the$_{\!}$ \textit{cluster$_{\!}$ center}$_{\!}$ matrix and$_{\!}$ $\bm{c}_{k\!}\!\in\!\mathbb{R}^{D{\!}}$ is$_{\!}$ $k^{th\!\!}$ cluster$_{\!}$ center;$_{\!\!}$ $\bm{M}\!\!=\!\![\bm{m}_1;\cdots\!;\bm{m}_N]^\top\!\!\in\!\!\mathbb{R}^{K\!\times\!N\!}$
is$_{\!}$ the$_{\!}$ \textit{cluster$_{\!}$ assignment}$_{\!}$ matrix$_{\!}$ and$_{\!}$ $\bm{m}_{n\!}\!\in_{\!}\!\{0,1\}^{K\!}$ is$_{\!}$ the$_{\!}$ one-hot$_{\!}$ assignment$_{\!}$ vector$_{\!}$ of$_{\!}$ $\bm{x}_n$;$_{\!}$ $\bm{1}_{K\!}$~is$_{\!}$~a$_{\!}$
$K_{_{\!\!}}$-dimensional all-ones$_{\!}$ vector.$_{\!}$ Principally,$_{\!}$ EM$_{\!}$ clustering$_{\!}$ works$_{\!}$ as$_{\!}$ follows:\\
{\ding{172}}$_{\!}$ \textbf{Cluster$_{\!}$ center$_{\!}$ initialization}:$_{\!}$ EM$_{\!}$ clustering$_{\!}$ starts$_{\!}$ with initial$_{\!}$ estimates$_{\!}$ for$_{\!}$ $K_{\!}$ cluster$_{\!}$ centers$_{\!}$ $\bm{C}^{(0)}\!=\![\bm{c}^{(0)}_1;\cdots\!;\bm{c}^{(0)}_K]$.  \\
{\ding{173}}$_{\!}$ \textbf{Iterative$_{\!}$ clustering$_{\!}$ and$_{\!}$ center$_{\!}$ update}: EM$_{\!}$ clustering proceeds by alternating between two steps:
\begin{itemize}[leftmargin=*]
		\setlength{\itemsep}{0pt}
		\setlength{\parsep}{-2pt}
		\setlength{\parskip}{-0pt}
		\setlength{\leftmargin}{-10pt}
		\vspace{-12pt}
\item \textit{Clustering (Expectation) Step} ``softly'' assigns each data samples to the $K$ clusters:
    \vspace{-2pt}
\begin{equation}\label{eq:kmeansupdate1}
	\begin{aligned}
	\hat{\bm{M}}^{(t)} = \mathop{\mathrm{softmax}}\nolimits_K (\bm{C}^{(t)}\bm{X}^{\top})\in[0,1]^{K\times N},
	\end{aligned}
\vspace{-3pt}
\end{equation}
where $\hat{\bm{M}}^{(t)\!}$ denotes the clustering probability matrix.
\item \textit{Update$_{\!}$ (Maximization)$_{\!}$ Step}$_{\!}$ recalculate$_{\!}$ each$_{\!}$ cluster$_{\!}$ center from$_{\!}$ the$_{\!}$ data$_{\!}$ according$_{\!}$ to$_{\!}$ their$_{\!}$ membership$_{\!}$ weights:
    \vspace{-5pt}
\begin{equation}\label{eq:kmeansupdate2}
	\begin{aligned}
	~~\bm{C}^{(t+1)} = \hat{\bm{M}}^{(t)\!}\bm{X}\in\mathbb{R}^{K\times D}.
	\end{aligned}
    \vspace{-7pt}
\end{equation}
	\end{itemize}
Apparently, the `'hard'' sample-to-cluster assignment can be given as: $\bm{M}\!=\!\mathrm{one}\text{-}\mathrm{hot}(\argmax\nolimits_{\!K\!}(\hat{\bm{M}}))$.
\vspace{-4pt}

\noindent\textbf{Cross-Attention for Clustering.} Inspired by DETR \cite{carion2020end}, recent end-to-end panoptic systems~\cite{wang2021max,zhang2021k,cheng2021per,li2022panoptic} are build upon a query-based scheme: a set~of $K$ queries $\bm{C}\!=\![\bm{c}_1;\cdots\!;\bm{c}_K]\!\in\!\mathbb{R}^{K\!\times\!D_{\!}}$ are learned and updated by a stack of transformer decoders for mask decoding. Here ``$\bm{C}$''$_{\!}$ is$_{\!}$ reused;$_{\!}$ we$_{\!}$ will$_{\!}$ relate$_{\!}$ queries$_{\!}$ with$_{\!}$ cluster$_{\!}$ centers$_{\!}$ later. Specifically, at each decoder, the cross-attention is adopted to adaptively aggregate pixel features to update the queries:
\vspace{-4pt}
\begin{equation}\label{eq:queryupdate}
	\begin{aligned}
	{\bm{C}} \leftarrow \bm{C} + \mathop{\mathrm{softmax}}\nolimits_{HW}(\bm{Q}^C(\bm{K}^I)^\top)\bm{V}^I,
	\end{aligned}
    \vspace{-4pt}
\end{equation}
where$_{\!}$ $\bm{Q}^{C\!\!}\!\in\!\mathbb{R}^{K\!\times\!D\!}, \bm{V}^{I\!\!}\!\in\!\mathbb{R}^{HW\!\times\!D\!}, \bm{K}^{I\!\!}\!\in\!\mathbb{R}^{HW\!\times\!D\!}$ are linearly projected$_{\!}$ features$_{\!}$ for$_{\!}$ query,$_{\!}$ key,$_{\!}$ and$_{\!}$ value;$_{\!}$ superscripts$_{\!}$ ``$C$'' and$_{\!}$ ``$I$''$_{\!}$ indicate$_{\!}$ the$_{\!}$ feature$_{\!}$ projected$_{\!}$ from$_{\!}$ the$_{\!}$ query$_{\!}$ and$_{\!}$ im- age features, respectively.
Inspired by \cite{yu2022cmt,yu2022k}, we reinterpret the cross-attention as a clustering solver by treating queries~as cluster centers, and applying \textit{softmax} on the query dimension ($K$) instead of image resolution ($HW$):
\vspace{-5pt}
\begin{equation}\label{eq:kmeansattention}
	\begin{aligned}
	{\bm{C}} \leftarrow \bm{C} + \mathop{\mathrm{softmax}}\nolimits_K(\bm{Q}^C(\bm{K}^I)^\top)\bm{V}^I.
	\end{aligned}
    \vspace{-2pt}
\end{equation}

\begin{figure*}[t]
	\begin{center}
		\includegraphics[width=\linewidth]{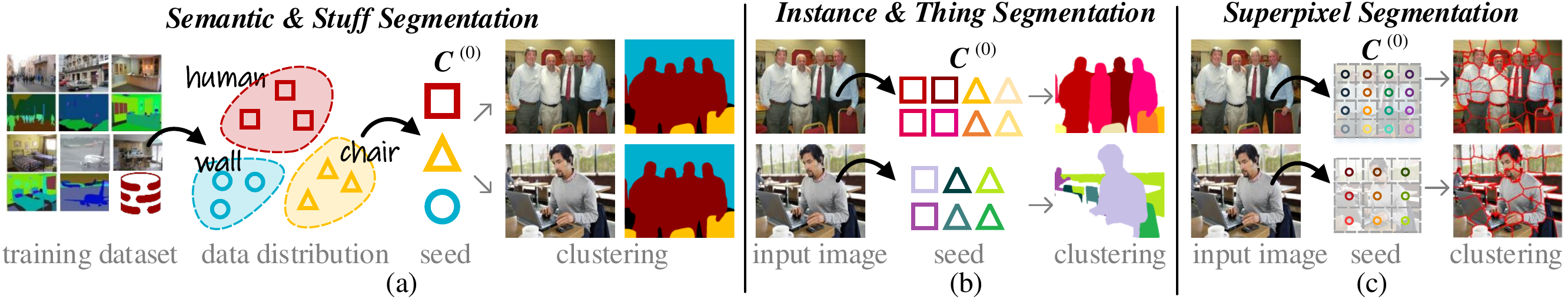}
	\end{center}
	\vspace{-15pt}
	\captionsetup{font=small}
	\caption{\small \textit{Dreamy-Start}$_{\!}$ for$_{\!}$ query$_{\!}$ initialization.$_{\!}$ (a)$_{\!}$ To$_{\!}$ respect$_{\!}$ the$_{\!}$ cross-scene$_{\!}$ semantically$_{\!}$ consistent$_{\!}$ nature$_{\!}$ of$_{\!}$ semantic/stuff$_{\!}$ segmentation, the$_{\!}$ quries/seeds$_{\!}$ are$_{\!}$ initialized$_{\!}$ as$_{\!}$ class$_{\!}$ centers$_{\!}$ (Eq.$_{\!}$~\ref{eq:stuffquery}).$_{\!}$ (b)$_{\!}$ To$_{\!}$ meet$_{\!}$ the$_{\!}$ instance-aware$_{\!}$ demand$_{\!}$ of$_{\!}$ instance/thing$_{\!}$ segmentation,$_{\!}$ the$_{\!}$ initial$_{\!}$ seeds are$_{\!}$ emerged$_{\!}$ from$_{\!}$ the$_{\!}$ input$_{\!}$ image$_{\!}$ (Eq.$_{\!}$~\ref{eq:thingquery}). (c)$_{\!}$ To$_{\!}$ generate$_{\!}$ varying$_{\!}$ number$_{\!}$ of$_{\!}$ superpixels,$_{\!}$ the$_{\!}$ seeds$_{\!}$ are$_{\!}$ initialized$_{\!}$ from$_{\!}$ image$_{\!}$ grids$_{\!}$ (Eq.$_{\!}$~\ref{eq:superpixelquery}).  }
	\vspace{-10pt}
	\label{fig:2}
\end{figure*}

\vspace{-5pt}
\subsection{\textsc{ClustSeg}}\label{sec:3.2}
\vspace{-2pt}
\textsc{ClustSeg}$_{\!}$ is$_{\!}$ built$_{\!}$ on$_{\!}$ the$_{\!}$ principle$_{\!}$ of$_{\!}$ segment-by-clustering: the$_{\!}$ segment$_{\!}$ masks$_{\!}$ $\{M_k\}_{k\!}$ in$_{\!}$ Eq.$_{\!}$~\ref{eq:1}$_{\!}$ correspond$_{\!}$ to$_{\!}$ the$_{\!}$ clustering assignment matrix $\bm{M}$ in Eq.~\ref{eq:kmeans}. Clustering can be further solved in a cross-attention form: the pixel-query affinities $\bm{Q}^{C\!}(\bm{K}^I)^{\!\top\!}$ in$_{\!}$ Eq.$_{\!}$~\ref{eq:kmeansattention}$_{\!}$ correspond$_{\!}$ to$_{\!}$ the$_{\!}$ clustering$_{\!}$ assignment probabilities$_{\!}$  $\bm{C}^{(t)\!}\bm{X}^{\!\top\!\!}$ in$_{\!}$ Eq.$_{\!}$~\ref{eq:kmeansupdate1}.$_{\!}$ In$_{\!}$ addition,$_{\!}$ with$_{\!}$ the$_{\!}$ close$_{\!}$ look at$_{\!}$ EM$_{\!}$ clustering$_{\!}$ (\textit{cf}.$_{\!}$~{\ding{172}}{\ding{173}}$_{\!}$ in$_{\!}$ \S\ref{sec:3.1}),$_{\!}$ two$_{\!}$ inherent$_{\!}$ defects$_{\!}$ of existing$_{\!}$ query-based$_{\!}$ segmentation$_{\!}$ models$_{\!}$ can$_{\!}$ be$_{\!}$ identified:
\begin{itemize}[leftmargin=*]
		\setlength{\itemsep}{0pt}
		\setlength{\parsep}{-2pt}
		\setlength{\parskip}{-0pt}
		\setlength{\leftmargin}{-10pt}
		\vspace{-10pt}
\item Due$_{\!}$ to$_{\!}$ the$_{\!}$ stochastic$_{\!}$ nature,$_{\!}$ EM$_{\!}$ clustering$_{\!}$ is$_{\!}$ highly$_{\!}$ sensi- tive$_{\!}$
     to$_{\!}$~the$_{\!}$ selection$_{\!}$ of$_{\!}$ initial$_{\!}$ centers$_{\!}$ (\textit{cf}.$_{\!}$~{\ding{172}})~\cite{celebi2013comparative}. To~alleviate the effects of \textit{initial starting conditions}, many initialization$_{\!}$ methods$_{\!}$ such$_{\!}$ as$_{\!}$ Forgy$_{\!}$ (randomly$_{\!}$ choose $K$~data samples as the initial centers)$_{\!}$~\cite{hamerly2002alternatives}$_{\!}$ are$_{\!}$ proposed.$_{\!}$ However,$_{\!}$ existing$_{\!}$ segmenters simply learn queries/centers in a \textit{fully parametric} manner, without any particular procedure of center initialization.

\item EM$_{\!}$ clustering$_{\!}$ provably$_{\!}$ converges$_{\!}$ to$_{\!}$ a$_{\!}$ local$_{\!}$ optimum$_{\!}$~\cite{vattani2009k}.$_{\!}$ However,$_{\!}$ it$_{\!}$ needs$_{\!}$ a$_{\!}$ sufficient$_{\!}$ number$_{\!}$ of$_{\!}$ itera- tions$_{\!}$ to$_{\!}$ do$_{\!}$ so$_{\!}$ (\textit{cf}.$_{\!}$~{\ding{173}}).$_{\!}$ Considering$_{\!}$ the$_{\!}$ computational$_{\!}$ cost and$_{\!}$ model$_{\!}$ size,$_{\!}$ existing$_{\!}$ segmenters$_{\!}$ only$_{\!}$ employ$_{\!}$ a$_{\!}$ few cross-attention based decoders$_{\!}$ (typically$_{\!}$ 6$_{\!}$ \cite{cheng2021per,yu2022cmt,yu2022k}),$_{\!}$ which$_{\!}$ may$_{\!}$ not$_{\!}$ enough$_{\!}$ to$_{\!}$ en- sure$_{\!}$ convergence$_{\!}$  from$_{\!}$ the$_{\!}$ perspective$_{\!}$ of$_{\!}$ EM$_{\!}$ clustering.
     		\vspace{-11pt}
	\end{itemize}

As a universal segmentation architecture, \textsc{ClustSeg} har- nesses$_{\!}$ the$_{\!}$ power$_{\!}$ of$_{\!}$ recursive$_{\!}$ clustering$_{\!}$ to$_{\!}$ boost$_{\!}$ pixel$_{\!}$ group- ing.$_{\!}$ It$_{\!}$ offers$_{\!}$ two$_{\!}$ innovative$_{\!}$ designs$_{\!}$ to$_{\!}$ respectively$_{\!}$ address the$_{\!}$~two$_{\!}$ defects:$_{\!}$ \textbf{i)}$_{\!}$ a$_{\!}$ well-crafted$_{\!}$ query-initialization$_{\!}$ scheme ---~\textit{dreamy-start} --- for$_{\!}$ the$_{\!}$ creation$_{\!}$ of$_{\!}$ informative$_{\!}$ initial$_{\!}$ clus- ter~centers; and \textbf{ii)} a non-parametric recursive module ---~\textit{re- current cross-attention} --- for effective neural clustering.
   		\vspace{-2pt}

Let $\bm{I}\!\in\!\mathbb{R}^{H_{\!}W_{\!}\times{\!}D_{\!\!}}$ denote the set of $D$-dimensional pixel em- beddings$_{\!}$ of$_{\!}$ image$_{\!}$ $I$.$_{\!}$ Analogous$_{\!}$ to$_{\!}$ EM$_{\!}$ clustering,$_{\!}$ \textsc{Clust- Seg}$_{\!}$ first$_{\!}$ creates$_{\!}$ a$_{\!}$ set$_{\!}$ of$_{\!}$ $K_{\!}$ queries$_{\!}$ $\bm{C}^{(0)\!}\!=\![\bm{c}^{(0)}_1;\cdots\!;\bm{c}^{(0)}_K]_{\!}$~as initial cluster centers using \textit{dreamy-start}. Then, \textsc{ClustSeg} iteratively conducts pixel clustering for mask decoding, by feeding$_{\!}$ pixel$_{\!}$ embeddings$_{\!}$ $\bm{I}_{\!}$ and$_{\!}$ the$_{\!}$ initial$_{\!}$ seeds$_{\!}$ $\bm{C}^{(0)\!}$ into$_{\!}$ a$_{\!}$ stack$_{\!}$ of \textit{recurrent cross-attention} decoders.
\vspace{-3pt}

\noindent\textbf{\textit{Dreamy-Start}$_{\!}$ for$_{\!}$ Query$_{\!}$ Initialization.$_{\!}$} \textit{Dreamy-start$_{\!}$} takes into$_{\!}$ account$_{\!}$ the$_{\!}$ heterogenous$_{\!}$ characteristics$_{\!}$ of$_{\!}$  different$_{\!}$ seg- mentation$_{\!}$ tasks$_{\!}$ for$_{\!}$ the$_{\!}$ creation$_{\!}$ of$_{\!}$ initial$_{\!}$ seeds$_{\!}$ $\bm{C}^{(0)\!}$ (Fig.$_{\!}$~\ref{fig:2}):
\vspace{-3pt}

\ding{228}$_{\!}$ \textit{Semantic$_{\!}$ Segmentation}$_{\!}$ groups$_{\!}$ pixels$_{\!}$ according$_{\!}$ to$_{\!}$ \textit{scene-/ instance-agnostic}$_{\!}$ semantic$_{\!}$ relations.$_{\!}$ For$_{\!}$ example,$_{\!}$ all$_{\!}$ the pixels of {dog}s should be grouped (segmented) into the same cluster,$_{\!}$ \ie,$_{\!}$ \textit{dog}$_{\!}$ class,$_{\!}$ regardless$_{\!}$ of$_{\!}$ whether$_{\!}$ they$_{\!}$ are$_{\!}$ from$_{\!}$ dif- ferent$_{\!}$ images/dog$_{\!}$ instances.$_{\!}$ Hence,$_{\!}$ for$_{\!}$ semantic-aware$_{\!}$ pixel$_{\!}$ clustering, the variance among different instances/scenes should$_{\!}$ be$_{\!}$ ignored.$_{\!}$ In$_{\!}$ this$_{\!}$ regard,$_{\!}$ we$_{\!}$ explore$_{\!}$ global$_{\!}$ semantic structures$_{\!}$ of$_{\!}$ the$_{\!}$ entire$_{\!}$ dataset$_{\!}$ to$_{\!}$ find$_{\!}$ robust$_{\!}$ initial$_{\!}$ seeds.$_{\!}$ Specifically, during training, we build a memory bank $\mathcal{B}_{\!}$~to store$_{\!}$ massive$_{\!}$ pixel$_{\!}$ samples$_{\!}$ for$_{\!}$ approximating$_{\!}$ the$_{\!}$ global data distribution. $\mathcal{B}_{\!}$ consists$_{\!}$ of$_{\!}$ $K_{\!}$ fixed-size,$_{\!}$ first-in-first-out queues, \ie, $\mathcal{B}\!=\!\{\mathcal{B}_1, \cdots\!, \mathcal{B}_{K}\}$; $\mathcal{B}_k$ stores numerous pixel embeddings which are sampled from training images and belong to class $k$. The initial query for cluster (class) $k$ is given as the corresponding ``class center'':
\vspace{-1pt}
\begin{equation}\label{eq:stuffquery}
	\begin{aligned} [\bm{c}^{(0)}_1;\cdots\!;\bm{c}^{(0)}_{K}]&=\mathrm{FFN}([\bar{\bm{x}}_1;\cdots\!;\bar{\bm{x}}_{K}]), \\ \bar{\bm{x}}_{k} &= \mathrm{Avg\_Pool}(\mathcal{B}_{k})\in\mathbb{R}^D,
	\end{aligned}
    \vspace{-1pt}
\end{equation}
where$_{\!}$ $\mathrm{Avg\_Pool}$ indicates average pooling, $\mathrm{FFN}$ is a fully-connected feed-forward network, and $K$ is set as the size of semantic vocabulary. In this way, the initial centers explicitly summarize the global statistics of the classes, facilitating scene-agnostic semantic relation based pixel clustering. Once trained, these initial seeds will be preserved for testing.
\vspace{-16pt}

\begin{figure*}[t]
	\begin{center}
		\includegraphics[width=\linewidth]{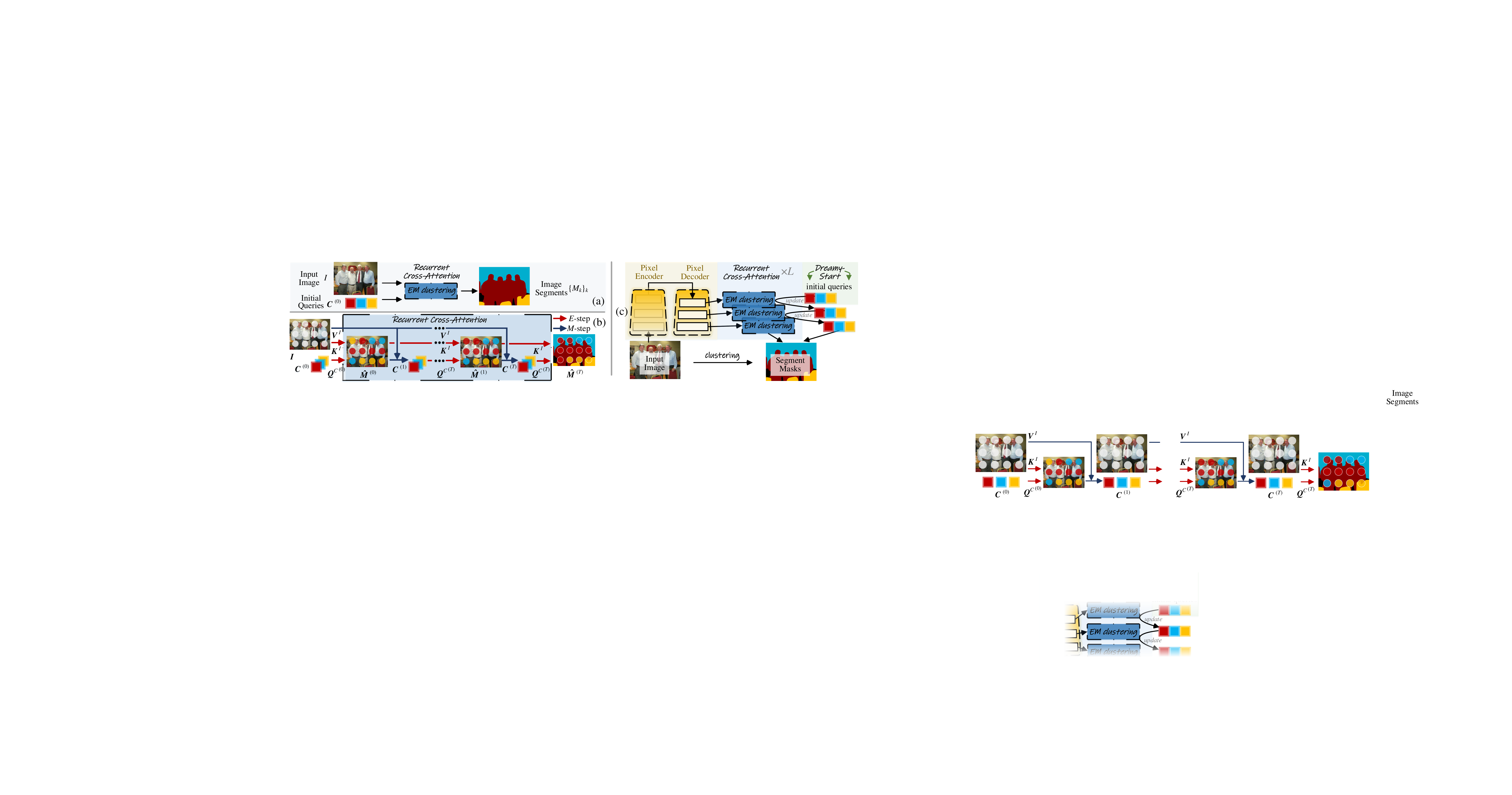}
	\end{center}
	\captionsetup{font=small}
	\vspace{-10pt}
	\caption{\small (a) \textit{Recurrent Cross-attention} instantiates EM clustering for segment-by-clustering. (b) Each \textit{Recurrent Cross-attention} layer executes $T$ iterations of clustering assignment (\textit{E}-step) and center update (\textit{M}-step). (c) Overall architecture of \textsc{ClustSeg}. }
	\vspace{-10pt}
	\label{fig:3}
\end{figure*}

\ding{228}$_{\!}$ \textit{Instance$_{\!}$ Segmentation}$_{\!}$ groups$_{\!}$ pixels$_{\!}$ according$_{\!}$ to$_{\!}$ \textit{instance-aware}$_{\!}$ relations$_{\!}$ ---$_{\!}$ pixels$_{\!}$ of$_{\!}$ different$_{\!}$ dog$_{\!}$ instances$_{\!}$ should$_{\!}$ be clustered into different groups. Different instances possess distinctive properties, \eg, color, scale, position, that are concerning with the local context --- the images --- that the instances situated in. It is hard to use a small finite set of $K$ fixed queries to characterize all the possible instances. Therefore, unlike previous methods learning $K$ changeless queries for different images, we derive our initial guess of instance-aware centers in an image context-adaptive manner:
\vspace{-5pt}
\begin{equation}\label{eq:thingquery}
	\begin{aligned}
	[\bm{c}^{(0)}_1;\cdots\!;\bm{c}^{(0)}_{K}] =\mathrm{FFN}(\mathrm{PE}(\bm{I})),
	\end{aligned}
    \vspace{-0pt}
\end{equation}
where$_{\!}$ $\mathrm{PE}_{\!}$ denotes$_{\!}$ position$_{\!}$ embedding, and $K$ is set as a constant (\ie, 100) --- much larger than the typical number of object instances in an image. As such, we utilize image-specific appearance and position cues to estimate content-adaptive seeds for instance-relation-oriented pixel grouping.
\vspace{-16pt}

\ding{228}$_{\!}$ \textit{Panoptic$_{\!}$ Segmentation} groups stuff and thing pixels in terms of semantic and instance relations respectively. Thus we respectively adopt the initialization strategies for semantic segmentation and instance segmentation to estimate two discrete sets of queries for stuff and thing pixel clustering.
\vspace{-2pt}

\ding{228}$_{\!}$ \textit{Superpixel$_{\!}$ Segmentation}$_{\!}$ groups$_{\!}$ together$_{\!}$ pixels$_{\!}$ that$_{\!}$ are spatially$_{\!}$ close$_{\!}$ and$_{\!}$ perceptually$_{\!}$ similar.$_{\!}$ The$_{\!}$ number$_{\!}$ of$_{\!}$ su- perpixels$_{\!}$ $K_{\!}$ is$_{\!}$ manually$_{\!}$ specified$_{\!}$ beforehand$_{\!}$ and$_{\!}$ can$_{\!}$ be$_{\!}$~ar- bitrary. We thus initialize the queries from image grids:
\vspace{-2pt}
\begin{equation}\label{eq:superpixelquery}
	\begin{aligned}
	[\bm{c}^{(0)}_1;\cdots\!;\bm{c}^{(0)}_{K}] =\mathrm{FFN}(\mathrm{Grid\_Sample}_K(\mathrm{PE}(\bm{I}))),
	\end{aligned}
    \vspace{-2pt}
\end{equation}
where $\mathrm{Grid\_Sample}_K(\mathrm{PE}(\bm{I}))$ refers to select $K$ position-embedded pixel features from $\mathrm{PE}(\bm{I})$ using grid-based sampling. The queries are used to group their surrounding pixels as superpixels. \textsc{ClustSeg} is thus general enough to accommodate the classic idea of grid-based clustering~\cite{achanta2012slic,yang2020superpixel} in superpixel segmentation.
\vspace{-2pt}

\textit{Dreamy-Start}$_{\!}$ renders$_{\!}$ \textsc{ClustSeg}$_{\!}$ with$_{\!}$ great$_{\!}$ flexibility$_{\!}$ of addressing$_{\!}$ task-specific$_{\!}$ properties$_{\!}$ without$_{\!}$ changing$_{\!}$ network architecture. Through customized initialization, high-quality cluster seeds are created for better pixel grouping.

\noindent\textbf{\textit{Recurrent Cross-Attention}$_{\!}$ for$_{\!}$ Recursive$_{\!}$ Clustering.$_{\!}$} After \textit{Dreamy-Start} based cluster center initialization, \textsc{ClustSeg} groups image pixels into $K$ clusters for segmentation, by resembling the workflow of EM clustering (\textit{cf}.$_{\!}$~{\ding{172}}{\ding{173}}$_{\!}$ in$_{\!}$ \S\ref{sec:3.1}).\\
Given$_{\!}$ the$_{\!}$ pixel$_{\!}$ embeddings$_{\!}$ $\bm{I}\!\in\!\mathbb{R}^{H_{\!}W_{\!}\times{\!}D_{\!\!}}$ and$_{\!}$ initial$_{\!}$ centers$_{\!}$ $\bm{C}^{(0)\!}$,$_{\!}$ the$_{\!}$ iterative$_{\!}$ procedure$_{\!}$ of$_{\!}$ EM$_{\!}$ clustering$_{\!}$ with$_{\!}$ $T_{\!}$ itera- tions~is encapsulated into a \textit{Recurrent Cross-Attention} layer:
\vspace{-8pt}
\begin{equation}\label{eq:recurrent}
	\begin{aligned}\small
\!\!\!\!\!\!\!\!\!E\text{-step:~~~~~} \hat{\bm{M}}^{(t)} &\!=\! \mathop{\mathrm{softmax}}\nolimits_K(\bm{Q}^{C^{(t)}}\!(\bm{K}^I)^\top),\!\!\!\!\!\!\!\!\\
\!\!\!\!\!\!\!\!\!M\text{-step:~~} \bm{C}^{(t+1)} &\!=\! \hat{\bm{M}}^{(t)}\bm{V}^I\!\in\!\mathbb{R}^{K\!\times\!D},
	\end{aligned}
    \vspace{-0pt}
\end{equation}
where$_{\!}$ $t_{\!}\!\in_{\!}\!\{1, \cdots\!, T\}$,$_{\!}$ and$_{\!}$ $\hat{\bm{M}}_{\!}\!\in_{\!}\![0,1]^{K\!\times\!HW\!\!}$ is$_{\!}$ the$_{\!}$ ``soft''$_{\!}$ clu- ster$_{\!}$ assignment$_{\!}$ matrix$_{\!}$ (\ie,$_{\!}$ probability$_{\!}$ maps$_{\!}$ of$_{\!}$ $K_{\!}$ segments). As$_{\!}$ defined$_{\!}$ in$_{\!}$ \S\ref{sec:3.1},$_{\!}$ $\bm{Q}^{C\!\!}\!\in\!\mathbb{R}^{K\!\times\!D\!}$ is$_{\!}$ the$_{\!}$ query$_{\!}$ vector$_{\!}$ projected from$_{\!}$ the$_{\!}$ center$_{\!}$ $\bm{C}$,$_{\!}$ and$_{\!}$ $\bm{V}^{I\!},_{\!} \bm{K}^{I\!\!}\!\in\!\mathbb{R}^{HW\!\times\!D\!}$ are$_{\!}$ the$_{\!}$ value$_{\!}$ and key$_{\!}$ vectors$_{\!}$  respectively$_{\!}$ projected$_{\!}$ from$_{\!}$ the$_{\!}$ image$_{\!}$ pixel$_{\!}$ fea- tures$_{\!}$ $\bm{I}$.$_{\!}$ \textit{Recurrent$_{\!}$ Cross-Attention}$_{\!}$ iteratively$_{\!}$ updates$_{\!}$ cluster$_{\!}$ membership $\hat{\bm{M}}$ (\ie, $E$-step) and centers $\bm{C}$ (\ie, $M$-step). It enjoys a few appealing characteristics (Fig.~\ref{fig:3}(a-b)):
\begin{itemize}[leftmargin=*]
		\setlength{\itemsep}{0pt}
		\setlength{\parsep}{-2pt}
		\setlength{\parskip}{-0pt}
		\setlength{\leftmargin}{-10pt}
		\vspace{-10pt}
\item \textit{Efficient:}$_{\!}$ Compared$_{\!}$ to$_{\!}$ the$_{\!}$ vanilla$_{\!}$ cross-attention$_{\!}$ (\textit{cf}.$_{\!}$~Eq.$_{\!}$~\ref{eq:queryupdate}) with$_{\!}$ the$_{\!}$ computational$_{\!}$ complexity$_{\!}$ $\mathcal{O}(H^2W^2D)$,$_{\!}$ \textit{Recur- rent$_{\!}$ Cross-Attention}$_{\!}$ is$_{\!}$ $\mathcal{O}(TKHWD)$,$_{\!}$ which$_{\!}$ is$_{\!}$ more$_{\!}$ effi- cient$_{\!}$ since$_{\!}$ $TK_{\!}\!\ll_{\!}\!HW$.$_{\!}$ Note$_{\!}$ that,$_{\!}$ during$_{\!}$ iteration,$_{\!}$ only$_{\!}$ $\bm{Q}$ needs$_{\!}$ to$_{\!}$ be$_{\!}$ recalculated,$_{\!}$ while$_{\!}$ $\bm{K}_{\!}$ and$_{\!}$ $\bm{V}_{\!}$ are$_{\!}$ only$_{\!}$ calculated once --- the small superscript $(t)$ is only added for $\bm{Q}$.
\item \textit{Non-parametric$_{\!}$ recursive:$_{\!}$} As$_{\!}$ the$_{\!}$ projection$_{\!}$ weights$_{\!}$ for query, key, and value are shared across iteration, \textit{Recurrent Cross-Attention} achieves recursiveness without occurring extra learnable parameters.
\item \textit{Transparent:} Aligning closely with the well-established EM clustering algorithm, \textit{Recurrent Cross-Attention} is crystal-clear and grants \textsc{ClustSeg} better transparency.
\item \textit{Effective:} \textit{Recurrent Cross-Attention} exploits the power of recursive clustering to progressively decipher the imagery intricacies. As a result, \textsc{ClustSeg} is more likely to converge to a better configuration of image partition.
    \vspace{-6pt}
\end{itemize}
We adopt a hierarchy of \textit{Recurrent Cross-Attention} based decoders to fully pursue the representational granularity for more effective pixel clustering:
\vspace{-2pt}
\begin{equation}\label{eq:decoder}
	\begin{aligned}
	{\bm{C}}^{l}\!=\!\bm{C}^{l+1\!}\!+\!\mathrm{RCross\_Attention}^{l+1}(\bm{I}^{l+1}, \bm{C}^{l+1}),
	\end{aligned}
    \vspace{-2pt}
\end{equation}
where $\bm{I}^{l\!}$ is the image feature map at $H/2^{l}\!\times\!W/2^{l}$ resolution,
and$_{\!}$ $\bm{C}^{l\!}$ is$_{\!}$ the$_{\!}$ cluster$_{\!}$ center$_{\!}$ matrix$_{\!}$ for$_{\!}$ $l^{th\!}$ decoder.$_{\!}$ The$_{\!}$~multi-head$_{\!}$ mechanism$_{\!}$
and$_{\!}$ multi-layer$_{\!}$ perceptron$_{\!}$ used$_{\!}$ in$_{\!}$ standard transformer$_{\!}$ decoder$_{\!}$ are$_{\!}$ also$_{\!}$ adopted$_{\!}$ (but$_{\!}$ omitted$_{\!}$ for$_{\!}$ simpli- city).$_{\!}$ The$_{\!}$ parameters$_{\!}$ for$_{\!}$ different$_{\!}$ \textit{Recurrent$_{\!}$ Cross-Attention} layers, \ie, $\{\mathrm{RCross\_Attention}^l\}^L_{l=1}$, are not shared.

\begin{table*}[!t]
    \vspace{-5pt}
    					\captionsetup{font=small}
					\caption{\small{Quantitative results} on COCO Panoptic~\cite{kirillov2019panoptic} \texttt{val} for \textbf{panoptic segmentation} (see  \S\ref{sec:PaS} for details). }
					\small
					\centering
					\resizebox{0.99\textwidth}{!}{
						\setlength\tabcolsep{6pt}
						\renewcommand\arraystretch{1.02}
						\begin{tabular}{r||l|c||ccccc }
							\thickhline
							\rowcolor{mygray}
							Algorithm & Backbone &Epoch & {PQ}$\uparrow$  &$\text{PQ}^\text{Th}$$\uparrow$ &$\text{PQ}^\text{St}$$\uparrow$ &$\text{AP}^\text{Th}_\text{pan}$$\uparrow$ &$\text{mIoU}_\text{pan}$$\uparrow$ \\
							
							\hline
							\hline
							
							Panoptic-FPN \cite{kirillov2019panopticfpn}
							& ResNet-101 & 20 & 44.0 & 52.0 & 31.9 & 34.0 & 51.5 \\
							UPSNet \cite{xiong2019upsnet}
							& ResNet-101 & 12 & 46.2 & 52.8 & 36.5 & 36.3 & 56.9 \\
							Panoptic-Deeplab \cite{cheng2020panoptic}
							& Xception-71 & 12 & 41.2 & 44.9 & 35.7 & 31.5 & 55.4 \\
							Panoptic-FCN \cite{li2021fully}
							& ResNet-50 & 12 & 44.3 & 50.0 & 35.6 & 35.5 & 55.0 \\
							Max-Deeplab \cite{wang2021max}
							& Max-L & 55 & 51.1 & 57.0 & 42.2 & --& -- \\
							CMT-Deeplab \cite{yu2022cmt}
							& Axial-R104$^{\dagger}$ & 55 &  54.1 & 58.8 & 47.1 & -- & -- \\
							\arrayrulecolor{gray}\hdashline\arrayrulecolor{black}
							
							\multirow{2}{*}{Panoptic Segformer \cite{li2022panoptic}}
							& ResNet-50 & \multirow{2}{*}{24} & 49.6$_{\textcolor{gray}{\pm0.25}}$ & 54.4$_{\textcolor{gray}{\pm0.26}}$ & 42.4$_{\textcolor{gray}{\pm0.25}}$ & 39.5$_{\textcolor{gray}{\pm0.20}}$ & 60.8$_{\textcolor{gray}{\pm0.21}}$ \\
							& ResNet-101 & & 50.6$_{\textcolor{gray}{\pm0.21}}$ & 55.5$_{\textcolor{gray}{\pm0.24}}$ & 43.2$_{\textcolor{gray}{\pm0.20}}$ & 40.4$_{\textcolor{gray}{\pm0.21}}$ & 62.0$_{\textcolor{gray}{\pm0.22}}$ \\
							\arrayrulecolor{gray}\hdashline\arrayrulecolor{black}
							
							\multirow{2}{*}{kMaX-Deeplab \cite{yu2022k}}
							&ResNet-50 & \multirow{3}{*}{50} & 52.1$_{\textcolor{gray}{\pm0.15}}$ & 57.3$_{\textcolor{gray}{\pm0.18}}$ & 44.0$_{\textcolor{gray}{\pm0.16}}$ & 36.2$_{\textcolor{gray}{\pm0.15}}$ & 60.4$_{\textcolor{gray}{\pm0.14}}$ \\
							&ConvNeXt-B$^{\dagger}$ &  & 56.2$_{\textcolor{gray}{\pm0.19}}$ & 62.4$_{\textcolor{gray}{\pm0.22}}$ & 46.8$_{\textcolor{gray}{\pm0.21}}$ & 42.2$_{\textcolor{gray}{\pm0.24}}$ & 65.3$_{\textcolor{gray}{\pm0.19}}$ \\
							
							\arrayrulecolor{gray}\hdashline\arrayrulecolor{black}	
							\multirow{2}{*}{K-Net \cite{zhang2021k}}
							& ResNet-101 & \multirow{2}{*}{36} & 48.4$_{\textcolor{gray}{\pm0.26}}$ & 53.3$_{\textcolor{gray}{\pm0.28}}$ & 40.9$_{\textcolor{gray}{\pm0.22}}$ & 38.5$_{\textcolor{gray}{\pm0.25}}$ & 60.1$_{\textcolor{gray}{\pm0.20}}$ \\
							& Swin-L$^{\dagger}$ &&55.2$_{\textcolor{gray}{\pm0.22}}$ & 61.2$_{\textcolor{gray}{\pm0.25}}$ & 46.2$_{\textcolor{gray}{\pm0.19}}$ & 45.8$_{\textcolor{gray}{\pm0.23}}$ & 64.4$_{\textcolor{gray}{\pm0.21}}$\\
							\arrayrulecolor{gray}\hdashline\arrayrulecolor{black}
							\multirow{3}{*}{Mask2Former \cite{cheng2021masked}}
							&ResNet-50 & \multirow{4}{*}{50} & 51.8$_{\textcolor{gray}{\pm0.24}}$ & 57.7$_{\textcolor{gray}{\pm0.23}}$ & 43.0$_{\textcolor{gray}{\pm0.16}}$ & 41.9$_{\textcolor{gray}{\pm0.23}}$ & 61.7$_{\textcolor{gray}{\pm0.20}}$ \\
							& ResNet-101 && 52.4$_{\textcolor{gray}{\pm0.22}}$ & 58.2$_{\textcolor{gray}{\pm0.16}}$ & 43.6$_{\textcolor{gray}{\pm0.22}}$ & 42.4$_{\textcolor{gray}{\pm0.20}}$ & 62.4$_{\textcolor{gray}{\pm0.21}}$\\
							& Swin-B$^{\dagger}$ && 56.3$_{\textcolor{gray}{\pm0.21}}$ & 62.5$_{\textcolor{gray}{\pm0.24}}$ & 46.9$_{\textcolor{gray}{\pm0.18}}$ & 46.3$_{\textcolor{gray}{\pm0.23}}$ & 65.1$_{\textcolor{gray}{\pm0.21}}$\\
							\hline
							\hline	
							\multirow{4}{*}{{\textbf{\textsc{ClustSeg}}} (ours)}
							&ResNet-50 & \multirow{4}{*}{50} & 54.3$_{\textcolor{gray}{\pm0.20}}$ & 60.4$_{\textcolor{gray}{\pm0.22}}$ & 45.8$_{\textcolor{gray}{\pm0.23}}$ & 42.2$_{\textcolor{gray}{\pm0.18}}$ & 63.8$_{\textcolor{gray}{\pm0.25}}$ \\
							& ResNet-101 && 55.3$_{\textcolor{gray}{\pm0.21}}$ & 61.3$_{\textcolor{gray}{\pm0.15}}$ & 46.4$_{\textcolor{gray}{\pm0.17}}$ & 43.0$_{\textcolor{gray}{\pm0.19}}$ & 64.1$_{\textcolor{gray}{\pm0.25}}$ \\
							& ConvNeXt-B$^{\dagger}$ & & 58.8$_{\textcolor{gray}{\pm0.18}}$ & 64.5$_{\textcolor{gray}{\pm0.16}}$ & \textbf{48.8}$_{\textcolor{gray}{\pm0.22}}$ & 46.9$_{\textcolor{gray}{\pm0.17}}$ & \textbf{66.3}$_{\textcolor{gray}{\pm0.20}}$ \\
							& Swin-B$^{\dagger}$ && \textbf{59.0}$_{\textcolor{gray}{\pm0.20}}$ & \textbf{64.9}$_{\textcolor{gray}{\pm0.23}}$ & 48.7$_{\textcolor{gray}{\pm0.19}}$ & \textbf{47.1}$_{\textcolor{gray}{\pm0.21}}$& 66.2$_{\textcolor{gray}{\pm0.18}}$ \\\hline
							\multicolumn{8}{l}{$^\dagger$: backbone pre-trained on ImageNet-22K~\cite{deng2009imagenet}; the marker is applicable to other tables.}
					\end{tabular}}
					\vspace{-10pt}
					\label{main_result_pano}
				\end{table*}

    \vspace{-3pt}
\subsection{Implementation Details}\label{sec:arch}
    \vspace{-2pt}
\noindent\textbf{Detailed$_{\!}$ Architecture.$_{\!}$} \textsc{ClustSeg}$_{\!}$ has$_{\!}$ four$_{\!}$ parts$_{\!}$ (Fig.$_{\!}$~\ref{fig:3}(c)):$_{\!\!\!}$
\begin{itemize}[leftmargin=*]
	\setlength{\itemsep}{0pt}
	\setlength{\parsep}{-2pt}
	\setlength{\parskip}{-0pt}
	\setlength{\leftmargin}{-8pt}
	\vspace{-10pt}
	\item \textit{Pixel Encoder} extracts multi-scale dense representations $\{\bm{I}_l\}_l$ for image $I$. In \S\ref{sec:experiments}, we test \textsc{ClustSeg} on various CNN-based and vision-transformer backbones.
	\item \textit{Pixel Decoder}, placed on the top of the encoder, gradually recovers finer representations. As in~\cite{yu2022k,yu2022cmt,cheng2021per}, we use six axial blocks \cite{wang2020axial}, one at $L^{th\!}$ level and five at $(L\!-\!1)^{th\!}$ level. 
	\item \textit{Recurrent$_{\!}$ Cross-Attention$_{\!}$ based$_{\!}$ Decoder}$_{\!}$ performs$_{\!}$ iterative clustering for pixel grouping. Each \textit{Recurrent Cross-Attention} layer conducts three iterations of clustering, \ie, $T\!=\!3$, and six decoders are used: each two is applied~to the pixel  decoder at levels $L\!-\!2$, $L\!-\!1$ and $L$, respectively.
\item \textit{Dreamy-Start} creates informative initial centers for the first \textit{Recurrent Cross-attention} based decoder and is customized to different tasks. For semantic segmentation~and stuff classes in panoptic segmentation, the seeds are computed from the memory bank during training (\textit{cf}.$_{\!}$~Eq.$_{\!}$~\ref{eq:stuffquery}) and stored unchanged once training finished. In other cases, the seeds are built on-the-fly (\textit{cf}.$_{\!}$~Eqs.$_{\!}$~\ref{eq:thingquery} and~\ref{eq:superpixelquery}).
        \vspace{-9pt}
\end{itemize}
    \vspace{-1pt}
\noindent\textbf{Loss$_{\!}$ Function.$_{\!}$} \textsc{ClustSeg}$_{\!}$ can$_{\!}$ be$_{\!}$ applied$_{\!}$ to$_{\!}$ the$_{\!}$ four$_{\!}$ seg- mentation$_{\!}$ tasks,$_{\!}$ without$_{\!}$ architecture$_{\!}$ change.$_{\!}$ \textbf{We$_{\!}$ opt$_{\!}$ the standard$_{\!}$ loss$_{\!}$ design$_{\!}$ in$_{\!}$ each$_{\!}$ task$_{\!}$ setting$_{\!}$ for$_{\!}$ training$_{\!}$ (de- tails$_{\!}$ in$_{\!}$ the$_{\!}$ supplementary)}.$_{\!}$ In$_{\!}$ addition,$_{\!}$ recall$_{\!}$ that$_{\!}$ \textit{Recurrent Cross-attention} estimates the cluster probability matrix $\hat{\bm{M}}^{(t)\!}$ at$_{\!}$ each$_{\!}$ $E$-step$_{\!}$ (\textit{cf}.$_{\!}$~Eq.$_{\!\!}$~\ref{eq:recurrent});$_{\!}$ $\hat{\bm{M}}^{(t)\!}$ can$_{\!}$ be$_{\!}$ viewed$_{\!}$ as$_{\!}$ logit maps of $K$ segments.$_{\!}$ Therefore,$_{\!}$ the$_{\!}$ groundtruth$_{\!}$ segment masks$_{\!}$ $\{M_k\}_{k\!}$ can$_{\!}$ be$_{\!}$ directly$_{\!}$ used$_{\!}$ to$_{\!}$ train$_{\!}$ every$_{\!}$ $E$-step$_{\!}$ of each \textit{Recurrent Cross-attention},$_{\!}$ leading$_{\!}$ to$_{\!}$ \textit{intermediate/deep~supervision}~\cite{lee2015deeply,yu2022k}


    \vspace{-7pt}
\section{Experiment}
\label{sec:experiments}
    \vspace{-3pt}
\textsc{ClustSeg} is the first framework to support four core~seg- mentation$_{\!}$ tasks$_{\!}$ with a$_{\!}$ single$_{\!}$ unified$_{\!}$ architecture.$_{\!}$ To$_{\!}$ demon- strate its broad applicability and wide benefit, we conduct

\begin{table*}[t]
					\captionsetup{font=small}
					\caption{\small{Quantitative results} on COCO~\cite{lin2014microsoft} \texttt{test-dev} for \textbf{instance segmentation} (see \S\ref{sec:InS} for details).}
					\centering
					\resizebox{0.99\textwidth}{!}{
						\setlength\tabcolsep{5pt}
						\renewcommand\arraystretch{1.02}
						\begin{tabular}{r||l|c||cc cc c c }
							\thickhline
							\rowcolor{mygray}
							Algorithm & Backbone &Epoch &{AP}$\uparrow$ &$\text{AP}_{50}$$\uparrow$ &$\text{AP}_{75}$$\uparrow$ &$\text{AP}_S$ &$\text{AP}_M$$\uparrow$ &$\text{AP}_L$$\uparrow$\\
							\hline
							\hline
							Mask R-CNN \cite{he2017mask}
							& ResNet-101 & 12 & 36.1 & 57.5 & 38.6 & 18.8 & 39.7 & 49.5 \\
							Cascade MR-CNN \cite{cai2019cascade}
							& ResNet-101 & 12 & 37.3 & 58.2 & 40.1 & 19.7 & 40.6 & 51.5 \\
							HTC \cite{chen2019hybrid}
							& ResNet-101 & 20 & 39.6 & 61.0 & 42.8 & 21.3 & 42.9 & 55.0 \\
							PointRend \cite{kirillov2020pointrend}
							& ResNet-50 & 12 & 36.3 & 56.9 & 38.7 & 19.8 & 39.4 & 48.5 \\
							BlendMask \cite{chen2020blendmask}
							& ResNet-101 & 36 & 38.4 & 60.7 & 41.3 & 18.2 & 41.5 & 53.3  \\
							QueryInst \cite{fang2021instances}
							&ResNet-101 & 36 & 41.0 & 63.3 & 44.5 & 21.7 & 44.4 & 60.7 \\
							
							SOLQ \cite{dong2021solq}
							& Swin-L$^{\dagger}$ & 50 & 46.7 & 72.7 & 50.6 & 29.2 & 50.1 & 60.9 \\
							SparseInst \cite{cheng2022sparse}
							&ResNet-50&36&37.9&59.2&40.2&15.7&39.4&56.9 \\
							\arrayrulecolor{gray}\hdashline\arrayrulecolor{black}

							\multirow{2}{*}{kMaX-Deeplab \cite{yu2022k}}
							& ResNet-50 & \multirow{2}{*}{50} & 40.2$_{\textcolor{gray}{\pm0.19}}$ & 61.5$_{\textcolor{gray}{\pm0.20}}$ & 43.7$_{\textcolor{gray}{\pm0.18}}$ & 21.7$_{\textcolor{gray}{\pm0.21}}$ & 43.0$_{\textcolor{gray}{\pm0.19}}$ & 54.0$_{\textcolor{gray}{\pm0.22}}$ \\
							&ConvNeXt-B$^{\dagger}$ &  & 44.7$_{\textcolor{gray}{\pm0.24}}$ & 67.5$_{\textcolor{gray}{\pm0.25}}$ & 48.1$_{\textcolor{gray}{\pm0.21}}$ & 25.1$_{\textcolor{gray}{\pm0.17}}$ & 47.6$_{\textcolor{gray}{\pm0.23}}$ & 61.5$_{\textcolor{gray}{\pm0.21}}$ \\

							\arrayrulecolor{gray}\hdashline\arrayrulecolor{black}	
							\multirow{2}{*}{K-Net \cite{zhang2021k}} & ResNet-101 & \multirow{2}{*}{36} &      40.1$_{\textcolor{gray}{\pm0.17}}$ & 62.8$_{\textcolor{gray}{\pm0.23}}$ & 43.1$_{\textcolor{gray}{\pm0.19}}$ & 18.7$_{\textcolor{gray}{\pm0.22}}$ & 42.7$_{\textcolor{gray}{\pm0.18}}$ & 58.8$_{\textcolor{gray}{\pm0.20}}$ \\
							& Swin-L$^{\dagger}$ && 46.1$_{\textcolor{gray}{\pm0.18}}$ & 67.7$_{\textcolor{gray}{\pm0.20}}$ & 49.6$_{\textcolor{gray}{\pm0.19}}$ & 24.3$_{\textcolor{gray}{\pm0.23}}$ & 49.5$_{\textcolor{gray}{\pm0.21}}$ & 65.1$_{\textcolor{gray}{\pm0.23}}$             \\
							\arrayrulecolor{gray}\hdashline\arrayrulecolor{black}	
							
							\multirow{3}{*}{Mask2Former \cite{cheng2021masked}}
							& ResNet-50 & \multirow{3}{*}{50} & 42.8$_{\textcolor{gray}{\pm0.23}}$ & 65.3$_{\textcolor{gray}{\pm0.21}}$ & 46.0$_{\textcolor{gray}{\pm0.22}}$ & 22.1$_{\textcolor{gray}{\pm0.19}}$ & 46.3$_{\textcolor{gray}{\pm0.21}}$ & 64.8$_{\textcolor{gray}{\pm0.23}}$ \\
							& ResNet-101 && 43.9$_{\textcolor{gray}{\pm0.19}}$ & 66.7$_{\textcolor{gray}{\pm0.17}}$ & 47.0$_{\textcolor{gray}{\pm0.19}}$ & 22.9$_{\textcolor{gray}{\pm0.20}}$ & 47.7$_{\textcolor{gray}{\pm0.15}}$ & 66.3$_{\textcolor{gray}{\pm0.18}}$ \\
							& Swin-B$^{\dagger}$ && 47.9$_{\textcolor{gray}{\pm0.19}}$ & 68.9$_{\textcolor{gray}{\pm0.18}}$ & 51.8$_{\textcolor{gray}{\pm0.21}}$ & 29.9$_{\textcolor{gray}{\pm0.23}}$ & 51.5$_{\textcolor{gray}{\pm0.20}}$& 68.5$_{\textcolor{gray}{\pm0.18}}$ \\
							\hline
							\hline
							\multirow{4}{*}{{\textbf{\textsc{ClustSeg}}} (ours)}
							&ResNet-50 & \multirow{4}{*}{50} & 44.2$_{\textcolor{gray}{\pm0.25}}$ & 66.7$_{\textcolor{gray}{\pm0.27}}$ & 47.8$_{\textcolor{gray}{\pm0.24}}$ & 24.3$_{\textcolor{gray}{\pm0.20}}$ &48.5$_{\textcolor{gray}{\pm0.21}}$&64.3$_{\textcolor{gray}{\pm0.24}}$  \\
							& ResNet-101 && 45.5$_{\textcolor{gray}{\pm0.22}}$ & 67.8$_{\textcolor{gray}{\pm0.21}}$ & 48.9$_{\textcolor{gray}{\pm0.24}}$ & 25.1$_{\textcolor{gray}{\pm0.20}}$ & 50.3$_{\textcolor{gray}{\pm0.23}}$ & 66.9$_{\textcolor{gray}{\pm0.27}}$\\
							& ConvNeXt-B$^{\dagger}$ && 49.0$_{\textcolor{gray}{\pm0.23}}$ & \textbf{70.4}$_{\textcolor{gray}{\pm0.22}}$ & 52.7$_{\textcolor{gray}{\pm0.20}}$ & \textbf{30.1}$_{\textcolor{gray}{\pm0.18}}$ & 52.9$_{\textcolor{gray}{\pm0.24}}$ & \textbf{68.6}$_{\textcolor{gray}{\pm0.25}}$ \\
							& Swin-B$^{\dagger}$ && \textbf{49.1}$_{\textcolor{gray}{\pm0.21}}$ & 70.3$_{\textcolor{gray}{\pm0.20}}$ & \textbf{52.9}$_{\textcolor{gray}{\pm0.23}}$ & \textbf{30.1}$_{\textcolor{gray}{\pm0.18}}$ & \textbf{53.2}$_{\textcolor{gray}{\pm0.20}}$& 68.4$_{\textcolor{gray}{\pm0.21}}$ \\
							\hline
					\end{tabular}}
					\vspace{-6pt}
					\label{result_instance}
				\end{table*}

 \textbf{Extensive$_{\!}$ experiments}:$_{\!}$ We$_{\!}$ benchmark$_{\!}$ it$_{\!}$ on$_{\!}$ panoptic$_{\!}$ (\S\ref{sec:PaS}), instance$_{\!}$ (\S\ref{sec:InS}),$_{\!}$ semantic$_{\!}$ (\S\ref{sec:SeS}),$_{\!}$ and$_{\!}$ superpixel (\S\ref{sec:SuS}) seg- mentation,$_{\!}$ and$_{\!}$ carry$_{\!}$ out$_{\!}$ ablation$_{\!}$ study$_{\!}$ (\S\ref{sec:DE}).$_{\!}$ We$_{\!}$ also$_{\!}$ ap- proach$_{\!}$ it$_{\!}$ on$_{\!}$ \textbf{diverse$_{\!}$ backbones}:$_{\!}$ ResNet$_{\!}$~\cite{he2016deep}, ConvNeXt$_{\!}$~\cite{liu2022convnet}, and Swin$_{\!}$~\cite{liu2021swin}.

\vspace{-5pt}
\subsection{Experiment on Panoptic Segmentation}\label{sec:PaS}\vspace{-3pt}
\noindent\textbf{Dataset.$_{\!}$} We$_{\!}$ use$_{\!}$ COCO$_{\!}$ Panoptic$_{\!}$~\cite{kirillov2019panoptic}$_{\!}$ --- \texttt{train2017}$_{\!}$ is adopted for training and \texttt{val2017} for test.\\
\noindent\textbf{Training.$_{\!}$} We$_{\!}$ set$_{\!}$ the$_{\!}$ initial$_{\!}$ learning$_{\!}$ rate$_{\!}$ to$_{\!}$ 1e-5,$_{\!}$ training epoch$_{\!}$ to$_{\!}$ 50,$_{\!}$ and$_{\!}$ batch$_{\!}$ size$_{\!}$ to$_{\!}$ 16.$_{\!}$  We$_{\!}$ use$_{\!}$ random$_{\!}$ scale$_{\!}$ jitter- ing$_{\!}$ with$_{\!}$ a$_{\!}$ factor$_{\!}$ in$_{\!}$ $[0.1, 2.0]_{\!}$ and$_{\!}$ a$_{\!}$ crop$_{\!}$ size$_{\!}$ of$_{\!}$ $1024\!\times\!1024$.\\
\noindent\textbf{Test.} We use one input image scale with shorter side as 800.\\
\noindent\textbf{Metric.} We use PQ~\cite{kirillov2019panoptic} and also report {PQ}$^\text{Th}$ and {PQ}$^\text{St}$ for ``thing'' and ``stuff'' classes, respectively. For completeness, we involve {AP}$^\text{Th}_\text{pan}$, which is {AP} evaluated on ``thing'' classes using instance segmentation annotations, and {mIoU}$_\text{pan}$, which is {mIoU} for semantic segmentation by
merging instance masks from the same category, using the same model trained for panoptic segmentation task.\\
\noindent\textbf{Performance$_{\!}$ Comparison.$_{\!}$} We$_{\!}$ compare$_{\!}$ \textsc{ClustSeg}$_{\!}$ with$_{\!}$ two families$_{\!}$ of$_{\!}$ state-of-the-art$_{\!}$ methods:$_{\!}$ \textit{universal}$_{\!}$ approa- ches$_{\!}$ (\ie,$_{\!}$ K-Net$_{\!}$ \cite{zhang2021k}, Mask2Former$_{\!}$ \cite{cheng2021masked}), and \textit{specialized} panoptic systems \cite{kirillov2019panopticfpn,xiong2019upsnet,cheng2020panoptic,li2021fully,wang2021max,zhang2021k,li2022panoptic,yu2022cmt}. As shown in Table~\ref{main_result_pano}, \textsc{ClustSeg} beats all universal rivals, \ie, Mask2Former and K-Net,~on COCO$_{\!}$ Panoptic$_{\!}$ \texttt{val}.$_{\!\!}$ With$_{\!}$ ResNet-50/-101,$_{\!}$ \textsc{ClustSeg}$_{\!}$ out- performs Mask2Former by $\textbf{2.3\%$/$2.9\%}$ PQ; with$_{\!}$ Swin-B, the$_{\!}$ margin$_{\!}$ is$_{\!}$ {$\textbf{\textbf{2.7}\%}$}$_{\!}$ PQ.$_{\!}$ Also,$_{\!}$ \textsc{ClustSeg}'s$_{\!}$ performance is clearly ahead of K-Net (\textbf{59.0\%}$_{\!}$~\textit{vs.}$_{\!}$~55.2\%), even using a lighter backbone (Swin-B$_{\!}$~\textit{vs.}$_{\!}$~Swin-L). Furthermore, \textsc{ClustSeg} outperforms all the well-established specialist panoptic algorithms. Notably, it achieves promising gains of $\textbf{2.6\%}$/ $\textbf{2.1}_{\!}\%$/$\textbf{2.0}\%_{\!}$ in$_{\!}$ terms$_{\!}$ of$_{\!}$ PQ/PQ$^\text{th}$/PQ$^\text{St\!\!}$ against$_{\!}$ kMax-Deeplab$_{\!}$ on the top of ConvNeXt-B. Beyond metric PQ, \textsc{ClustSeg} gains$_{\!}$ superior$_{\!}$ performance$_{\!}$ in$_{\!}$ terms$_{\!}$ of$_{\!}$ AP$^\text{Th}_\text{pan\!}$ and$_{\!}$ mIoU$_{\text{pan}}$. In$_{\!}$ summary,$_{\!}$ {\textsc{ClustSeg}},$_{\!}$ with$_{\!}$ Swin-B$_{\!}$ backbone,$_{\!}$ \textbf{\textit{sets$_{\!}$ new records across all the metrics on COCO Panoptic \texttt{val}}}.

    \vspace{-6pt}
\subsection{Experiment on Instance Segmentation}\label{sec:InS}
    \vspace{-3pt}
\noindent\textbf{Dataset.$_{\!}$} As$_{\!}$ standard,$_{\!}$ we$_{\!}$ adopt$_{\!}$ COCO$_{\!}$~\cite{lin2014microsoft}$_{\!}$ --- \texttt{train2017} is used for training and \texttt{test-dev} for test.\\
\noindent\textbf{Training.$_{\!}$} We$_{\!}$ set$_{\!}$ the$_{\!}$ initial$_{\!}$ learning$_{\!}$ rate$_{\!}$ to$_{\!}$ 1e-5,$_{\!}$ training epoch$_{\!}$ to$_{\!}$ 50,$_{\!}$ and$_{\!}$ batch$_{\!}$ size$_{\!}$ to$_{\!}$ 16.$_{\!}$  We$_{\!}$ use$_{\!}$ random$_{\!}$ scale$_{\!}$ jitter- ing$_{\!}$ with$_{\!}$ a$_{\!}$ factor$_{\!}$ in$_{\!}$ $[0.1, 2.0]_{\!}$ and$_{\!}$ a$_{\!}$ crop$_{\!}$ size$_{\!}$ of$_{\!}$ $1024\!\times\!1024$.\\
\noindent\textbf{Test.} We use one input image scale with shorter side as 800.\\
\noindent\textbf{Metric.} We adopt {AP}, AP$_{50}$, {AP}$_{75}$, {AP}$_S$, {AP}$_M$, and {AP}$_L$.\\
\noindent\textbf{Performance$_{\!}$ Comparison.$_{\!}$} Table~\ref{result_instance} presents the results of \textsc{ClustSeg} against 11 famous instance segmentation methods on COCO \texttt{test-dev}. \textsc{ClustSeg} shows clear performance advantages over prior arts. With ResNet-101, it outperforms the universal counterparts Mask2Former by $\textbf{1.6\%}$ and K-Net by $\textbf{5.4\%}$ in terms of AP. It surpasses all the specialized competitors, \eg, yielding a significant gain of $\textbf{4.0\%}$ AP over kMax-Deeplab when using ResNet-50. Without bells and whistles, \textit{\textbf{\textsc{ClustSeg} establishes a new state-of-the-art on COCO instance segmentation}}.

\begin{table}[t]
\centering
\vspace{-8pt}
					\captionsetup{font=small}
					\caption{\small Quantitative results on ADE20K~\cite{zhou2017scene}~\texttt{val} for \textbf{semantic segmentation} (see \S\ref{sec:SeS} for details).}
\resizebox{0.5\textwidth}{!}{
\hspace{-1em}
						\setlength\tabcolsep{2pt}
						\renewcommand\arraystretch{1.02}
						\begin{tabular}{r||l|c||c}
							\thickhline
							\rowcolor{mygray}
							Algorithm & Backbone &Epoch & {mIoU}$\uparrow$ \\
							\hline
							\hline
							FCN \cite{long2015fully} & ResNet-50 &50& 36.0\\
							DeeplabV3+ \cite{deeplabv3plus2018}& ResNet-50 & 50 & 42.7\\
							APCNet \cite{he2019adaptive} & ResNet-50 & 100 & 43.4\\
							SETR \cite{zheng2021rethinking}&ViT-L$^{\dagger}$&100&49.3  \\
							Segmenter \cite{strudel2021segmenter}& ViT-L$^{\dagger}$& 100& 53.5 \\
							Segformer \cite{xie2021segformer}& MIT-B5 & 100 & 51.4  \\

							\arrayrulecolor{gray}\hdashline\arrayrulecolor{black}	
							
							\multirow{2}{*}{kMaX-Deeplab \cite{yu2022k}}&ResNet-50 & \multirow{2}{*}{100} & 48.1$_{\textcolor{gray}{\pm0.13}}$ \\
							& ConvNeXt-B$^{\dagger}$ & & 56.2$_{\textcolor{gray}{\pm0.16}}$ \\
							\arrayrulecolor{gray}\hdashline\arrayrulecolor{black}	
							\multirow{2}{*}{K-Net \cite{zhang2021k}}& ResNet-50 & \multirow{2}{*}{50} & 44.6	$_{\textcolor{gray}{\pm0.25}}$  \\
							& Swin-L$^{\dagger}$ && 53.7$_{\textcolor{gray}{\pm0.15}}$    \\
							\arrayrulecolor{gray}\hdashline\arrayrulecolor{black}	
							\multirow{2}{*}{Mask2Former \cite{cheng2021masked}}& ResNet-50 & \multirow{2}{*}{100} & 48.2$_{\textcolor{gray}{\pm0.12}}$  \\
							& Swin-B$^{\dagger}$ & & 54.5$_{\textcolor{gray}{\pm0.20}}$ \\
							\hline
							\hline
							\multirow{3}{*}{\textbf{\textsc{ClustSeg}} (ours)} &ResNet-50 & \multirow{3}{*}{100} & 50.5$_{\textcolor{gray}{\pm0.16}}$ \\
							& ConvNeXt-B$^{\dagger}$ & & 57.3$_{\textcolor{gray}{\pm0.17}}$ \\
							& Swin-B$^{\dagger}$ & & \textbf{57.4}$_{\textcolor{gray}{\pm0.22}}$ \\
							
							\hline
					\end{tabular}}
					\vspace{-10pt}
					\label{result_semantic}
				\end{table}

    \vspace{-6pt}
\subsection{Experiment on Semantic Segmentation}\label{sec:SeS}
    \vspace{-3pt}
\noindent\textbf{Dataset.} We experiment with {ADE20K}~\cite{zhou2017scene}, which includes 20K/2K/3K  images for \texttt{train}/\texttt{val}/\texttt{test}.\\
\noindent\textbf{Training.$_{\!}$} We$_{\!}$ set$_{\!}$ the$_{\!}$ initial$_{\!}$ learning$_{\!}$ rate$_{\!}$ to$_{\!}$ 1e-5,$_{\!}$ training epoch$_{\!}$ to$_{\!}$ 100,$_{\!}$ and$_{\!}$ batch$_{\!}$ size$_{\!}$ to$_{\!}$ 16.$_{\!}$  We$_{\!}$ use$_{\!}$ random$_{\!}$ scale$_{\!}$ jitter- ing with a factor in $[0.5, 2.0]$ and a crop size of $640\!\times\!640$.\\
\noindent\textbf{Test.} At the test time, we rescale the shorter side of input image to 640, without any test-time data augmentation.\\
\noindent\textbf{Metric.} Mean intersection-over-union (mIoU) is reported.\\
\noindent\textbf{Performance$_{\!}$ Comparison.$_{\!}$} In Table~\ref{result_semantic}, we further compare \textsc{ClustSeg} with a set of semantic segmentation methods on ADE20K \texttt{val}. \textsc{ClustSeg} yields superior performance. For$_{\!}$ example,$_{\!}$ it$_{\!}$ outperforms$_{\!}$ Mask2Former$_{\!}$ by$_{\!}$ $\textbf{2.3\%}_{\!}$~and$_{\!}$ $\textbf{2.9\%}$
mIoU using ResNet-50 and Swin-B$^\dagger$ backbones, respectively. Furthermore,  \textsc{ClustSeg} leads other specialist semantic segmentation models like Segformer \cite{xie2021segformer}, Segmenter \cite{strudel2021segmenter}, and SETR \cite{zheng2021rethinking} by large margins. Considering that ADE20K is challenging and extensively-studied, such improvements are particularly$_{\!}$ impressive.$_{\!}$ In$_{\!}$ conclusion,$_{\!}$ \textbf{\textit{\textsc{ClustSeg}$_{\!}$ ranks$_{\!}$ top$_{\!}$ $_{\!}$ in$_{\!}$ ADE20K$_{\!}$ semantic$_{\!}$ segmentation$_{\!}$ benchmarking}}.

\begin{figure}[t]
	\begin{center}
		\includegraphics[width=0.99\linewidth]{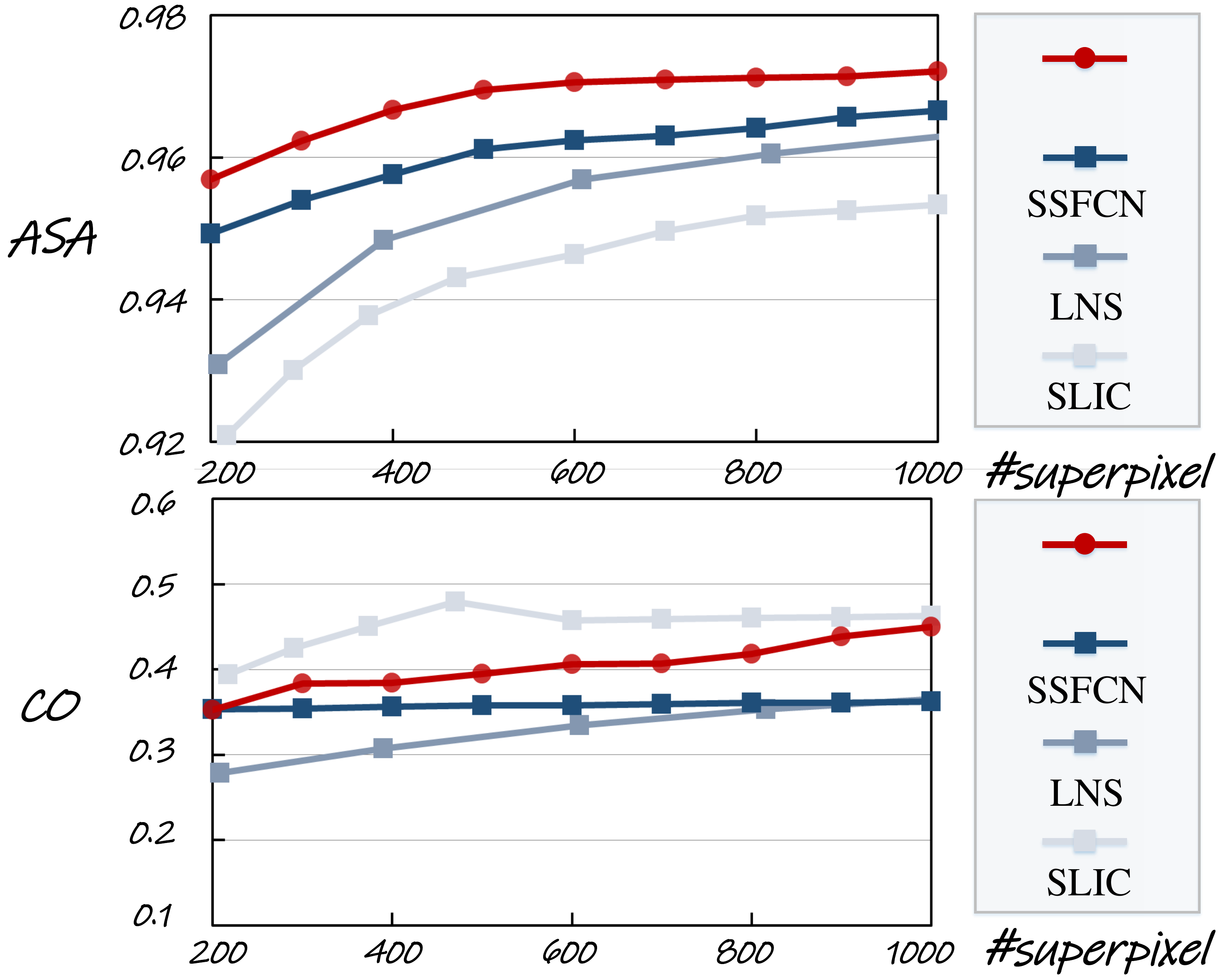}
    \put(-41,165){\scriptsize \textsc{ClustSeg} }
     \put(-41,73){\scriptsize \textsc{ClustSeg} }
	\end{center}
	\vspace{-14pt}
	\captionsetup{font=small}
	\caption{\small \textsc{ClustSeg} reaches the best ASA and CO scores on {BSDS500} \cite{arbelaez2011contour} \texttt{test}, among all the deep learning based superpixel  models (see \S\ref{sec:SuS} for details). }
	\vspace{-10pt}
	\label{fig:4}
\end{figure}

\begin{table*}[t]
					\caption{A set of \textbf{ablative studies} on COCO Panoptic~\cite{li2022panoptic} \texttt{val} (see \S\ref{sec:DE}). The adopted designs are marked in {\color{red}red}.}
					\begin{subtable}{0.4\linewidth}
						\captionsetup{width=.95\linewidth}
						\resizebox{\textwidth}{!}{
							\setlength\tabcolsep{4pt}
							\renewcommand\arraystretch{1.1}
							\begin{tabular}{l|ccc}
								\thickhline
								\rowcolor{mygray}
								Algorithm Component & PQ$\uparrow$  &$\text{PQ}^\text{Th\!\!}\uparrow$ &$\text{PQ}^\text{St\!\!}\uparrow$ \\
								\hline\hline
								\textsc{Baseline} & 49.7 & 55.5 & 42.0 \\
								\arrayrulecolor{gray}\hdashline\arrayrulecolor{black}	
								$+$ Dreamy-Start \textit{only}& 51.0 & 56.7 & 43.6 \\
								$+$ Recurrent Cross-attention \textit{only}	& 53.2 & 59.1 & 44.9 \\ 
								\arrayrulecolor{gray}\hdashline\arrayrulecolor{black}		
								\textbf{\textsc{Clustseg}} ({\color{red}\textit{both}}) & \textbf{54.3} & \textbf{60.4} & \textbf{45.8} \\
								\hline
						\end{tabular}}
						\setlength{\abovecaptionskip}{0.3cm}
						\setlength{\belowcaptionskip}{-0.1cm}
						\caption{\small{Key Component Analysis}}
						\label{table:components}
					\end{subtable}
                    \vspace{-6px}
					\hspace{-0.5em}\setcounter{subtable}{2}
					\begin{subtable}{0.6\linewidth}
						\resizebox{\textwidth}{!}{
							\setlength\tabcolsep{2pt}
							\renewcommand\arraystretch{1.09}
							\begin{tabular}{l|ccc|cc}
								\thickhline
								\rowcolor{mygray}
								Cross-Attention Variant & PQ$\uparrow$  &$\text{PQ}^\text{Th}$$\uparrow$ &$\text{PQ}^\text{St}$$\uparrow$ &\makecell{Training Speed\\{(hour/epoch)$\downarrow$}}&\makecell{Inference Speed \\{(fps)$\uparrow$}}\\
								\hline\hline
								Vanilla (Eq.$_{\!}$~\ref{eq:queryupdate}) & 51.0 & 56.7 & 43.6 & 1.89 & 5.88\\
								$K$-Means~\cite{yu2022k}  & 53.4 & 58.5 & 45.3 & 1.58 & 7.81\\
								{{\color{red}Recurrent}} (Eq.$_{\!}$~\ref{eq:decoder}) & \textbf{54.3} & \textbf{60.4} & \textbf{45.8} & 1.62 & 7.59\\
								\hline
							\end{tabular}
						}
						\setlength{\abovecaptionskip}{0.3cm}
						\setlength{\belowcaptionskip}{-0.1cm}
						\caption{\small{\textit{Recurrent Cross-Attention}}}
						\label{table:rca}
					\end{subtable}
					
					\begin{subtable}{0.54\linewidth}\setcounter{subtable}{1}
						\resizebox{\textwidth}{!}{
							\setlength\tabcolsep{4.5pt}
							\renewcommand\arraystretch{1.1}
							\begin{tabular}{c|cc|ccc|ccc}
								\thickhline
								\rowcolor{mygray}
								&\multicolumn{2}{c|}{\textit{Instance/Thing}} &  \multicolumn{3}{c|}{\textit{Semantic/Stuff}} & & & \\
								\rowcolor{mygray}
								\#&free &scene-&free &scene-&scene-& PQ$\uparrow$&$\text{PQ}^\text{Th}$$\uparrow$ &$\text{PQ}^\text{St}$$\uparrow$  \\
								\specialrule{0em}{-1.5pt}{-1.5pt}
								\rowcolor{mygray}
								&param.  &adaptive&param. &agnostic &adaptive &&& \\ \hline\hline
								1&\cmark& & \cmark & &   & 53.2 & 59.1 & 44.9 \\
								2&& \cmark & \cmark &  &    & 54.0 & 60.2 & 45.2\\
								3&\cmark & & & \cmark &  & 53.9 & 59.5 & 45.7 \\
								4&\cmark&  &  &  &\cmark  & 53.5 & 59.3 & 45.3 \\
								\arrayrulecolor{gray}\hdashline\arrayrulecolor{black}		
								5&& {\color{red}\cmark} & &{\color{red}\cmark}   &  & \textbf{54.3} & \textbf{60.4} & \textbf{45.8} \\
								\hline
							\end{tabular}
						}
						\setlength{\abovecaptionskip}{0.3cm}
						\setlength{\belowcaptionskip}{-0.1cm}
						\caption{\small{\textit{Dreamy-Start} Query-Initialization}}
						\label{table:ds}
					\end{subtable}
					\begin{subtable}{0.46\linewidth}\setcounter{subtable}{3}
						\resizebox{\textwidth}{!}{
							\setlength\tabcolsep{4.5pt}
							\renewcommand\arraystretch{1.07}
							\begin{tabular}{c|cc c|cc}
								\thickhline	\rowcolor{mygray}
								$T$ & PQ$\uparrow$  &$\text{PQ}^\text{Th}$$\uparrow$ &$\text{PQ}^\text{St}$$\uparrow$ &\makecell{Training Speed\\{(hour/epoch)$\downarrow$}}&\makecell{Inference Speed \\{(fps)$\uparrow$}}\\
								\hline
								\hline
								1 & 53.8 & 59.7 & 45.4 & 1.54 & 8.08\\
								2 & 54.1 & 60.2 & 45.7 & 1.59 & 7.85\\
								{\color{red}{3}} & 54.3 & 60.4 & 45.8 & 1.62 & 7.59 \\
								4 & 54.3 & 60.4 & 45.8 & 1.68 & 7.25\\
								5 & 54.3 & 60.5 & 45.8& 1.74 & 6.92\\
								6 & 54.4 & 60.4 & 45.9 & 1.82 & 6.54\\
								\hline
							\end{tabular}
						}
						\setlength{\abovecaptionskip}{0.3cm}
						\setlength{\belowcaptionskip}{-0.1cm}
						\caption{\small{Recursive Clustering}}
						\label{table:T}
					\end{subtable}
					\vspace{-30pt}
				\end{table*}

    \vspace{-5pt}
\subsection{Experiment on Superpixel Segmentation}\label{sec:SuS}
\vspace{-2pt}
\noindent\textbf{Dataset.} We use {BSDS500} \cite{arbelaez2011contour}, which includes 200/100/200 images for \texttt{train}/\texttt{val}/\texttt{test}.\\
\noindent\textbf{Training.$_{\!}$} We$_{\!}$ set$_{\!}$ the$_{\!}$ initial$_{\!}$ learning$_{\!}$ rate$_{\!}$ to$_{\!}$ 1e-4,$_{\!}$ training$_{\!}$ itera- tion$_{\!}$ to$_{\!}$ 300K,$_{\!}$ and$_{\!}$ batch$_{\!}$ size$_{\!}$ to$_{\!}$ 128.$_{\!}$  We$_{\!}$ use$_{\!}$ random horizontal and vertical flipping, random$_{\!}$ scale$_{\!}$ jittering with a factor in $[0.5, 2.0]$,  and a crop size of $480\!\times\!480$ for data augmentation. We randomly choose the number of superpixels from 50 to 2500. Note that the grid for query generation is automatically adjusted to match the specified number of superpixels.\\
\noindent\textbf{Test.} {During inference, we use the original image size.}\\
\noindent\textbf{Metric.} We use achievable segmentation accuracy (ASA) and compactness (CO). ASA is aware of boundary adherence, whereas CO addresses shape regularity. \\
\noindent\textbf{Performance$_{\!}$ Comparison.$_{\!}$} Fig~\ref{fig:4} presents comparison results of superpixel segmentation on BSDS500 \texttt{test}. In terms of ASA, \textsc{ClustSeg} outperforms  the classic method SLIC$_{\!}$~\cite{achanta2012slic} by a large margin, and also surpasses recent three deep learning based competitors, \ie,  SSFCN$_{\!}$~\cite{yang2020superpixel} and LNS$_{\!}$~\cite{zhu2021learning}. In addition, \textsc{ClustSeg} gains
high$_{\!}$ CO$_{\!}$ score.$_{\!}$ As$_{\!}$ seen,$_{\!}$ \textsc{ClustSeg}$_{\!}$ performs$_{\!}$ well$_{\!}$ on$_{\!}$ both ASA$_{\!}$ and$_{\!}$ CO;$_{\!}$ this$_{\!}$ is$_{\!}$ significant$_{\!}$ due$_{\!}$ to$_{\!}$ the$_{\!}$ well-known$_{\!}$~trade-off$_{\!}$ between$_{\!}$ edge-preserving$_{\!}$ and$_{\!}$ compactness$_{\!}$~\cite{yang2020superpixel}.$_{\!}$ Our$_{\!}$ \textbf{\textit{\textsc{ClustSeg}$_{\!}$ achieves$_{\!}$ outstanding$_{\!}$ performance against$_{\!}$ state-of-the-art$_{\!}$ superpixel$_{\!}$ methods$_{\!}$ on$_{\!}$ BSDS500}}.

    \vspace{-2pt}
\subsection{Diagnostic Experiment}\label{sec:DE}
    \vspace{-1pt}
In this section, we dive deep into \textsc{ClustSeg} by ablating of its key components on COCO Panoptic~\cite{kirillov2019panoptic} \texttt{val}. ResNet-50 is adopted as the  backbone. \textbf{More experimental
results are given in the supplementary}.

\noindent\textbf{Key$_{\!}$ Component$_{\!}$ Analysis.$_{\!}$} We$_{\!}$ first$_{\!}$ investigate$_{\!}$~the$_{\!}$ two$_{\!}$~ma- jor ingredients in \textsc{ClustSeg}, \ie, \textbf{\textit{Dreamy-Start}}~for query initialization and  \textbf{\textit{Recurrent Cross-Attention}},~for recursive clustering.$_{\!\!}$ We$_{\!}$ build$_{\!}$ \textsc{Baseline}$_{\!}$ that$_{\!}$ learns$_{\!}$ the$_{\!}$ initial queries fully end-to-end and updates them through standard cross-attention (Eq.$_{\!}$~\ref{eq:queryupdate}) based decoders.$_{\!}$ As reported in Table$_{\!}$~\ref{table:components},$_{\!}$ \textsc{Baseline}$_{\!}$ gives $49.7\%$ PQ, $55.5\%$ PQ$^\text{Th}$, and $42.0\%$ PQ$^\text{St}$. After applying \textit{Dreamy-Start} to \textsc{Baseline}, we observe consistent and notable improvements for both `thing' ($55.5\%_{\!}\rightarrow_{\!}\textbf{56.7\%}$ in PQ$^\text{Th}$) and `stuff' ($42.0\%_{\!}\rightarrow_{\!}\textbf{43.6\%}$ in PQ$^{\text{St}}$), leading to an increase of overall PQ  from $49.7\%$ to $\textbf{51.0\%}$. This reveals the critical role of object queries and verifies the efficacy of our query-initialization strategy, even without explicitly conducting clustering.  Moreover, after introducing \textit{Recurrent Cross-Attention} to \textsc{Baseline}, we obtain significant gains of $\textbf{3.5\%}$ PQ,  $\textbf{3.6\%}$ PQ$^\text{Th}$,  and $\textbf{2.9\%}$ PQ$^\text{St}$.  Last, by unifying the two core techniques together, \textsc{ClustSeg} yields the best performance across all the three metrics. This suggests that the proposed \textit{Dreamy-Start} and \textit{Recurrent Cross-Attention} can work collaboratively, and confirms the effectiveness of our overall algorithmic design.

\noindent \textbf{\textit{Dreamy-Start} Query-Initialization.} We next study the impact of the our \textit{Dreamy-Start} Query-Initialization scheme. As summarized in Table~\ref{table:ds}, when learning the initial queries as free parameters as standard (\#1), the model obtains $53.2\%$ PQ,$_{\!}$ $59.1\%_{\!}$ PQ$^\text{Th\!}$ and$_{\!}$ $44.9\%_{\!}$ PQ$^\text{St}$.$_{\!}$ By$_{\!}$ initializing$_{\!}$ `thing'$_{\!}$ centers in a scene context-adaptive manner (Eq.$_{\!}$~\ref{eq:thingquery}), we observe a large gain of $\textbf{1.1\%}$ PQ$^\text{Th}$ (\#2). Additionally, with scene-agnostic initialization of `stuff' centers (Eq.$_{\!}$~\ref{eq:stuffquery}), the model yields a clear boost of PQ$^\text{St}$ from $44.9\%$ to $\textbf{45.7\%}$ (\#3). In addition, we find that only minor gains are achieved for PQ$^\text{St}$ if `stuff' centers are also initialized as scene-adaptive (\#4). By customizing initialization strategies for both `thing' and `stuff' centers, \textit{Dreamy-Start} provides substantial performance improvements across all the metrics (\#5).

\noindent\textbf{\textit{Recurrent Cross-Attention}.} We further probe the influence of our \textit{Recurrent Cross-Attention} (Eq.~\ref{eq:decoder}), by comparing it with vanilla cross-attention (Eq.~\ref{eq:queryupdate}) and $K$-Means cross-attention$_{\!}$ \cite{yu2022k}.$_{\!}$ $K$-Means$_{\!}$ cross-attention$_{\!}$ employs Gumbel-Softmax \cite{jang2016categorical} for `hard' pixel-cluster assignment, without any recursive process. {As seen in Table~\ref{table:rca}, our \textit{Recurrent Cross-Attention} is \textit{effective} --- it improves the vanilla and $K$-Means$_{\!}$ by \textbf{3.3}\% PQ and \textbf{0.9}\% PQ respectively, and \textit{efficient} --- its training and inference speeds are much faster than the vanilla and comparable to $K$-Means, as consistent with our analysis in \S\ref{sec:3.2}.}

\noindent\textbf{Recursive$_{\!}$ Clustering.$_{\!}$} Last,$_{\!}$ to$_{\!}$ gain$_{\!}$ more$_{\!}$ insights$_{\!}$ into$_{\!}$ recur- sive clustering, we ablate the effect of iteration number~$T$ in$_{\!}$ Table$_{\!}$~\ref{table:T}.$_{\!}$ We$_{\!}$ find$_{\!}$ that$_{\!}$ the$_{\!}$ performance$_{\!}$ gradually improves from $53.8$\% PQ to $\textbf{54.3}$\% PQ when increasing $T$ from $1$ to $3$, but remains unchanged after running more iterations. Additionally, the speed of training and inference decreases as $T$ increases.$_{\!}$  We therefore set $T\!=\!3$ by default for a better trade-off between accuracy and computational cost.

    \vspace{-4pt}
\section{Conclusion}
\label{sec:conclusion}
    \vspace{-2pt}
In this work, our epistemology is centered on the \textit{segment-by-clustering} paradigm, which coins a universal framework, termed \textsc{ClustSeg}, to unify the community of image segmentation and respect the distinctive characteristics of each sub-task$_{\!}$ (\ie,$_{\!}$ superpixel,$_{\!}$ semantic,$_{\!}$ instance,$_{\!}$ and$_{\!}$ panoptic). The clustering insight leads us to introduce novel approaches for task-aware query/center initialization and tailor the cross-attention mechanism for recursive clustering. Empirical results suggest that \textsc{ClustSeg} achieves superior performance in all the four sub-tasks. Our research may potentially benefit the broader domain of dense visual prediction as a whole.

%
%


\bibliography{egbib}
\bibliographystyle{icml2023}

\newpage
\appendix
\onecolumn


In this document, we provide additional experimental results and analysis, pseudo code, more implementation details and discussions. It is organized as follows:
\begin{itemize}
    \item \S\ref{sec:detail}: More experimental details
    \item \S\ref{sec:AS}: More ablative studies
    \item \S\ref{sec:PC}: Pseudo code
    \item \S\ref{sec:dis}: More discussions

\end{itemize}

	\section{More Experimental Details}
	\label{sec:detail}
We provide more experimental results of \textsc{ClustSeg} (with Swin-B backbone) on five datasets: COCO
panoptic~\cite{kirillov2019panoptic} \texttt{val}  for \textbf{panoptic segmentation}
in Fig.~\ref{fig:pano}, COCO~\cite{lin2014microsoft} \texttt{val2017}  for \textbf{instance segmentation}  in Fig.~\ref{fig:ins}, ADE20K~\cite{zhou2017scene} \texttt{val} for
\textbf{semantic segmentation}
in Fig.~\ref{fig:sem}, and NYUv2~\cite{silberman2012indoor}  as well as BSDS500 \cite{arbelaez2011contour}  for \textbf{superpixel segmentation} in Fig.~\ref{fig:5} and Fig.~\ref{fig:sup}.
Our results demonstrate that \textsc{ClustSeg} can learn and discover, from the underlying characteristics of the data, the division principle of pixels, hence yielding strong performance across various core image segmentation tasks.
	
  \textbf{Implementation Details.} \textsc{ClustSeg} is implemented in PyTorch. All the~backbones$_{\!}$ are$_{\!}$ initialized$_{\!}$ using$_{\!}$ corresponding$_{\!}$ weights$_{\!}$ pre-trained on ImageNet-1K/-22K \cite{deng2009imagenet}, while the remaining layers are randomly initialized. We train all our models using AdamW optimizer and cosine annealing learning rate decay policy. For panoptic, instance, and semantic segmentation, we adopt the default training recipes of MMDetection \cite{mmdetection}.

	\subsection{Panoptic Segmentation}\label{sec:pano}
	
	\textbf{Dataset.} COCO panoptic~\cite{kirillov2019panoptic} is considered as a standard benchmark dataset in the field of panoptic segmentation, providing a rich and diverse set of images for training and evaluation. It is a highly advanced and sophisticated dataset that utilizes the full spectrum of annotated images from COCO~\cite{lin2014microsoft} dataset. COCO panoptic encompasses the 80 ``thing'' categories as well as an additional diligently annotated set of 53 ``stuff'' categories. To ensure the integrity and coherence of the dataset, any potential overlapping categories between the two aforementioned tasks are meticulously resolved. Following the practice of COCO, COCO Panoptic is divided into 115K/5K/20K images for \texttt{train}/ \texttt{val}/\texttt{test} split.

\textbf{Training.} Following \cite{carion2020end,max_deeplab_2021,yu2022k,cheng2021masked}, we set the total number of cluster seeds (\ie, queries) as 128, in which 75 are for ``thing'' and 53 are for ``stuff''. During training, we optimize the following objective:
	\begin{equation}\label{eq:loss}
		\mathcal{L}^\text{Panoptic} = \lambda^\text{th}\mathcal{L}^\text{th} + \lambda^\text{st}\mathcal{L}^\text{st} + \lambda^\text{aux}\mathcal{L}^\text{aux},
	\end{equation}
	where $\mathcal{L}^\text{th}$ and $\mathcal{L}^\text{st}$ are loss functions for things and stuff. For a fair comparison, we follow \cite{yu2022k,wang2021max} to additionally employ an auxiliary loss that is computed as a weighted summation of four loss terms, \ie, a PQ-style loss, a mask-ID cross-entropy loss, an instance discrimination loss, and a semantic segmentation loss. We refer to \cite{wang2021max,yu2022k} for more details about $\mathcal{L}^\text{aux}$. The coefficients $\lambda^\text{th}$, $\lambda^\text{st}$ and $\lambda^\text{aux}$ are set as: $\lambda^\text{th}=5$, $\lambda^\text{st}=3$, and $\lambda^\text{aux}=1$. In addition, the final ``thing'' centers are feed into a small FFN for semantic classification, trained with a binary cross-entropy loss.

\begin{figure*}[!ht]
    \centering
    \includegraphics[width=\textwidth]{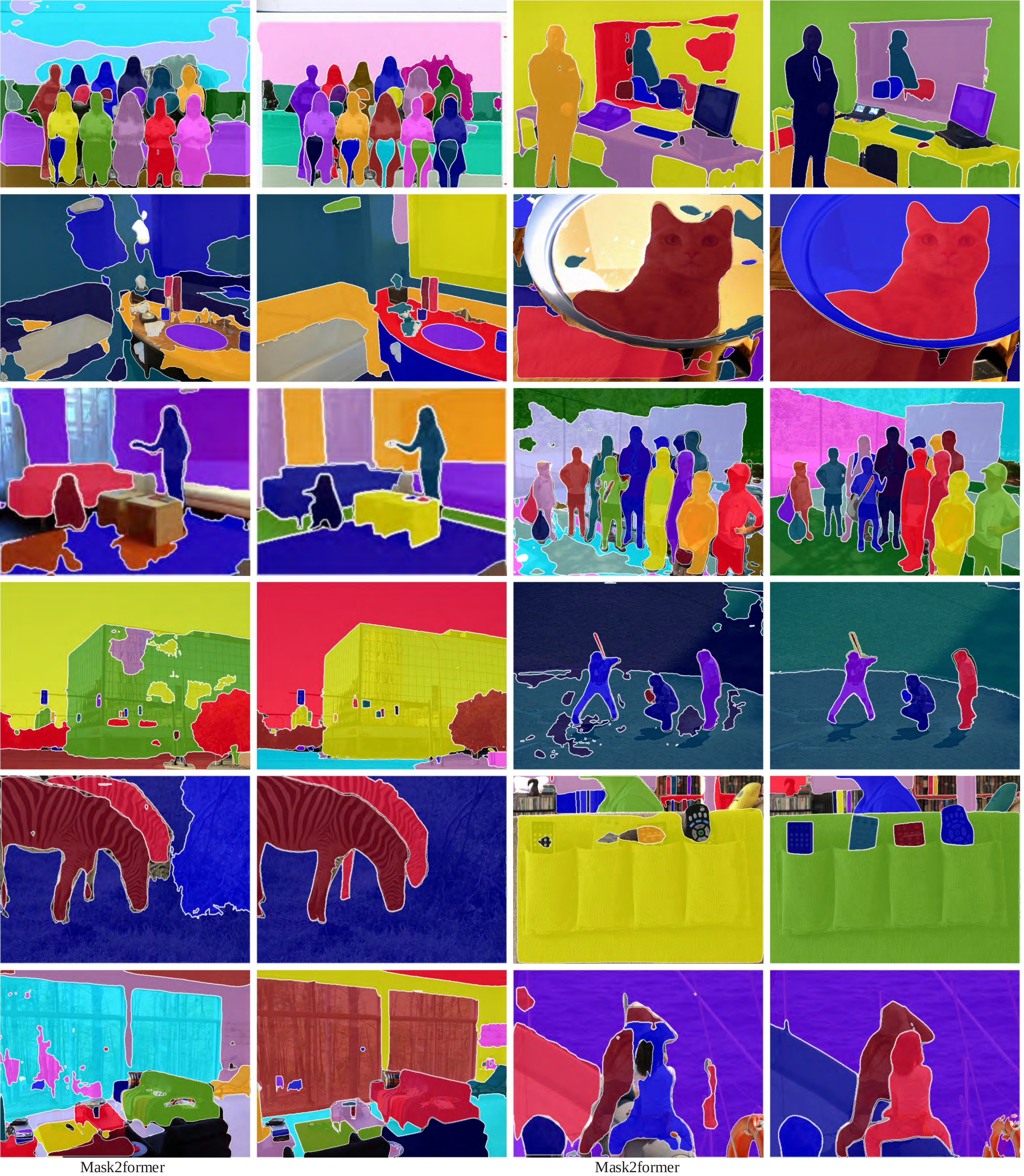}
         \put(-82,2){\scalebox{.8}{\textsc{ClustSeg}}}
     \put(-325,2){\scalebox{.8}{\textsc{ClustSeg}}}
    \caption{\textbf{Qualitative panoptic segmentation} results on COCO panoptic~\cite{kirillov2019panoptic} \texttt{val}. \textsc{ClustSeg} with Swin-B~\cite{liu2021swin} backbone achieves \textbf{59.0\%} \textit{PQ}.  See \S\ref{sec:pano} for details.}
    \label{fig:pano}
\end{figure*}

 \textbf{Qualitative Results.}  \textsc{ClustSeg} is capable to achieve appealing performance in various challenging scenarios. Specifically, in the \textit{restroom} example (see Fig.~\ref{fig:pano} row \#2 col \#1 and \#2), it perfectly segments the object instances and preserves more details of backgrounds within a highly intricate indoor scenario;  in the \textit{zebra} example (see Fig.~\ref{fig:pano} row \#5 col \#1 and \#2), \textsc{ClustSeg} successfully recognizes two distinct zebras with similar patterns as well as the grass backgrounds; in the \textit{person} example (see Fig.~\ref{fig:pano} row \#3 col \#3 and \#4), \textsc{ClustSeg} differentiates the person in the dense crowd and identifies the complex backgrounds.
	
	\subsection{Instance Segmentation} \label{sec:ins}
	
	\textbf{Dataset.} We use COCO~\cite{lin2014microsoft}, the golden-standard dataset for instance segmentation. It has dense annotations for 80 object categories, including common objects such as people, animals, furniture, vehicles.  The images in the dataset are diverse, covering a wide range of challenging indoor and outdoor scenes.  As standard, we use \texttt{train2017} split (115K images) for training, \texttt{val2017} (5K images) for validation, and \texttt{test-dev}  (20K images) for testing. All the results in the main paper are reported for \texttt{test-dev}.
	
	\textbf{Training.} For a fair comparison, we follow the training protocol in \cite{cheng2021masked}: 1)  the number of instance centers is set to 100; 2) a combination of the binary cross-entropy loss and the dice Loss is used as the optimization objective.  Their coefficients are set to $5$ and $2$, respectively. In addition, the final instance centers are feed into a small FFN for semantic classification, trained with a binary cross-entropy loss.

\begin{figure*}[!ht]
    \centering
    \includegraphics[width=\textwidth]{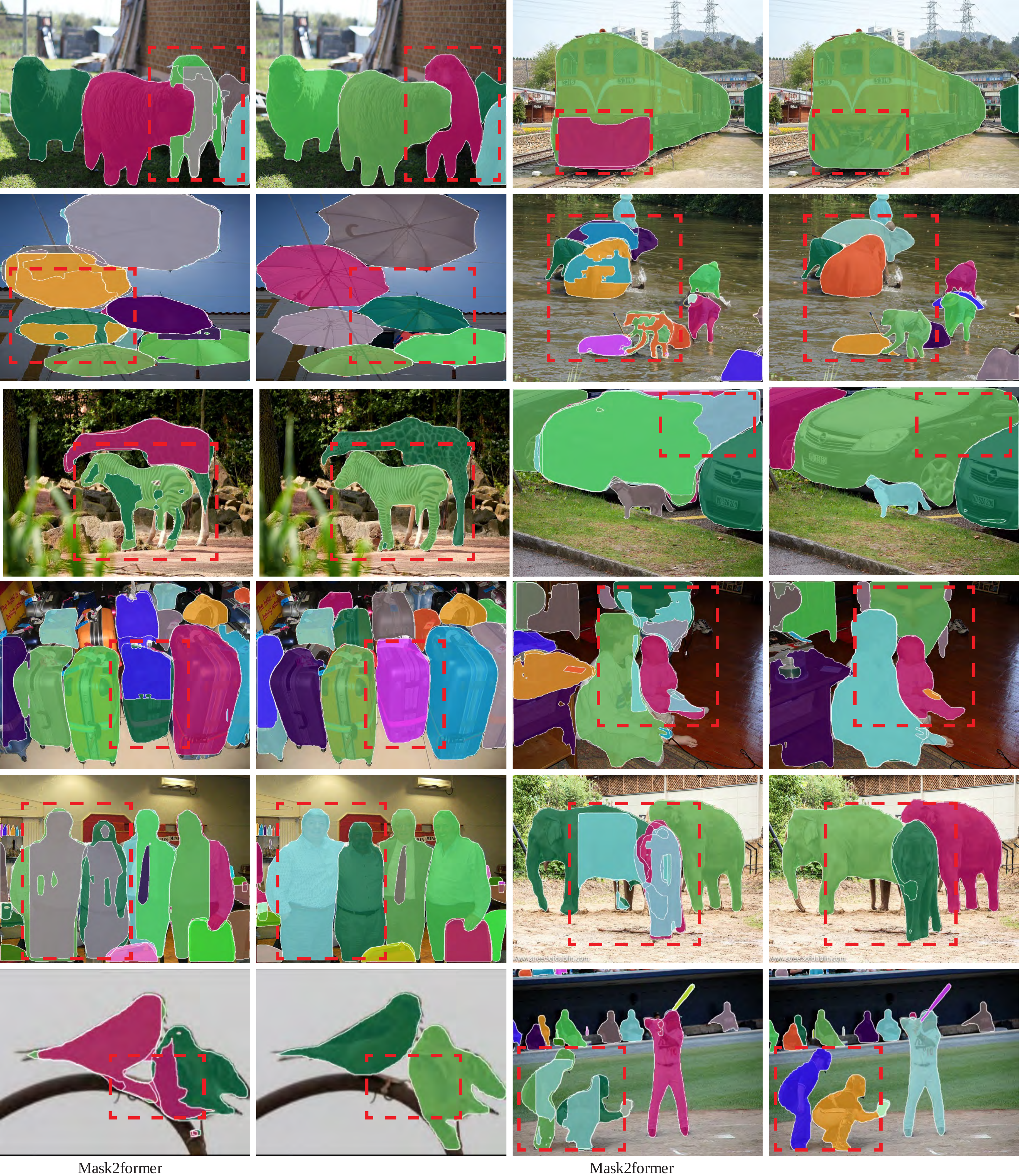}
         \put(-82,1){\scalebox{.8}{\textsc{ClustSeg}}}
     \put(-328,1){\scalebox{.8}{\textsc{ClustSeg}}}
    \caption{\textbf{Qualitative  instance segmentation} results on COCO~\cite{lin2014microsoft} \texttt{val2017}. \textsc{ClustSeg} with Swin-B~\cite{liu2021swin} backbone achieves \textbf{49.1\%} \textit{AP}.  See \S\ref{sec:ins} for details.}
    \label{fig:ins}
\end{figure*}

 \textbf{Qualitative Results.} Consistent to panoptic segmentation, \textsc{ClustSeg} also demonstrates strong efficacy in instance segmentation. For instance, in the \textit{elepants} example (see Fig.~\ref{fig:ins} row \#5 col \#3 and \#4), \textsc{ClustSeg} successfully separates apart a group of elephants under significant occlusions and similar appearance; in the \textit{river} example (see Fig.~\ref{fig:ins} row \#2 col \#3 and \#4), \textsc{ClustSeg} effectively distinguishes the highly-crowded and occluded person as well.
	
	\subsection{Semantic Segmentation} \label{sec:sem}
	
	\textbf{Dataset.}	{ADE20K}~\cite{zhou2017scene} is a large-scale scene parsing benchmark that covers a wide variety of indoor and outdoor scenes annotated with 150 semantic categories (\eg, door, cat, sky) . It is divided into 20K/2K/3K  images for \texttt{train}/\texttt{val}/\texttt{test}. The images cover many daily scenes, making it a challenging dataset for semantic segmentation.
	
	\textbf{Training.} In semantic segmentation, the number of cluster seeds is set to the number of semantic categories, \ie, 150 for ADE20K. We adopt the same loss function as  \cite{zhang2021k, cheng2021masked, strudel2021segmenter} by combining the standard cross-entropy loss with an auxiliary dice loss. By default, the coefficients for the two losses are set to $5$ and $1$, respectively.

\begin{figure*}[!ht]
    \centering
    \includegraphics[width=\textwidth]{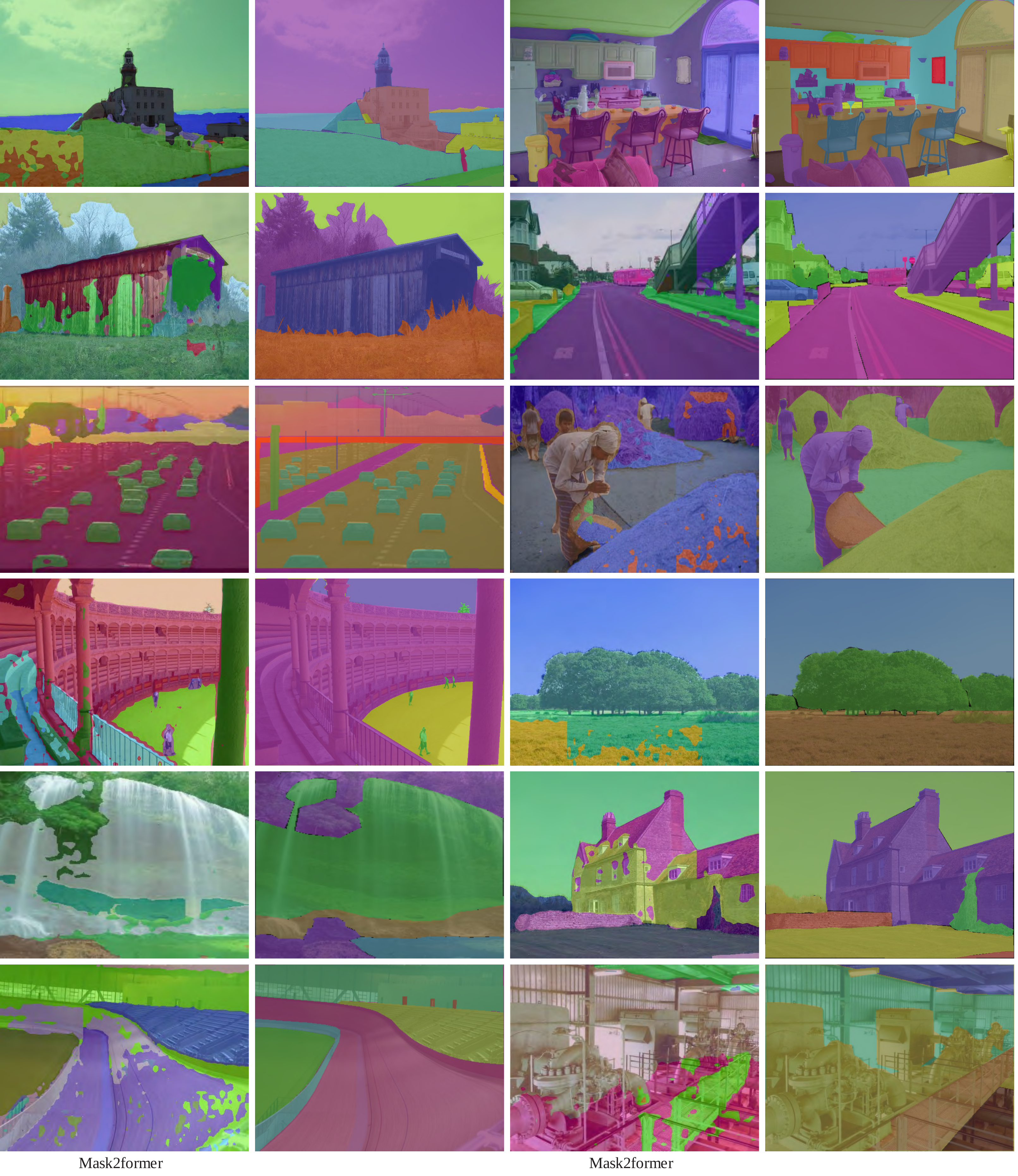}
         \put(-82,4.5){\scalebox{.8}{\textsc{ClustSeg}}}
     \put(-325,4.5){\scalebox{.8}{\textsc{ClustSeg}}}
    \caption{\textbf{Qualitative semantic segmentation results} on ADE20K~\cite{zhou2017scene} \texttt{val}. \textsc{ClustSeg} with Swin-B~\cite{liu2021swin} backbone achieves \textbf{57.4} \textit{mIoU}.  See \S\ref{sec:sem} for details.}
    \label{fig:sem}
\end{figure*}

\textbf{Qualitative Results.} When dealing with both indoor (see Fig.~\ref{fig:sem} row \#1 col \#3 and \#4) and outdoor (see Fig.~\ref{fig:sem} row \#2 col \#3 and \#4) scenarios, \textsc{ClustSeg} delivers highly accurate results. Especially, for the challenging outdoor settings.  \textsc{ClustSeg}  can robustly delineate the delicacy of physical complexity across the scenes, where Mask2Former, a recent top-leading segmentation algorithm,  generates a large array of wrongful mask predictions.

\begin{figure}[t]
	\begin{center}
		\includegraphics[width=0.99\linewidth]{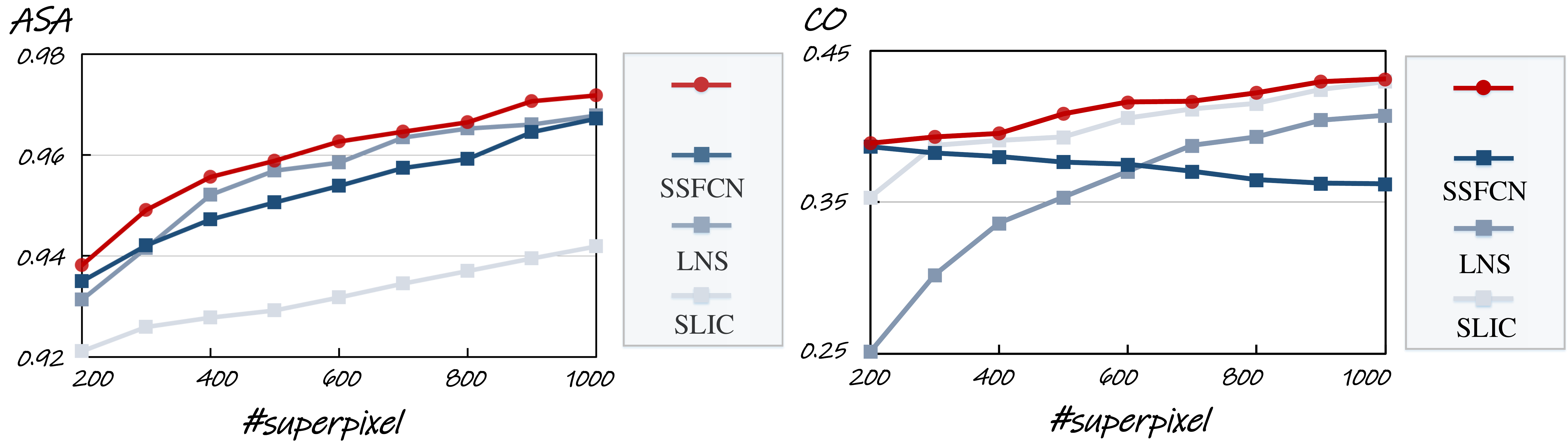}
    \put(-288,97){\footnotesize \textsc{ClustSeg} }
     \put(-46,97){\footnotesize \textsc{ClustSeg} }
	\end{center}
	\vspace{-14pt}
	\caption{ \textsc{ClustSeg} reaches the best ASA and CO scores on {NYUv2} \cite{silberman2012indoor} \texttt{test}  (see \S\ref{sec:sup} for details). }
	\vspace{-15pt}
	\label{fig:5}
\end{figure}

	\subsection{Superpixel Segmentation} \label{sec:sup}
	
	\textbf{Dataset.} For superpixel segmentation, we utilize two standard datasets (\ie, BSDS500~\cite{arbelaez2011contour} and NYUv2~\cite{silberman2012indoor}). {BSDS500} contains 500 natural images with pixel-wise semantic annotations. These image are divided into 200/100/200 for \texttt{train}/\texttt{val}/\texttt{test}. Following  \cite{yang2020superpixel,tu2018learning}, we train our model using the combination of all images in \texttt{train} and \texttt{val}, and run evaluation on \texttt{test}. NYUv2 dataset is originally proposed for indoor scene understanding tasks, which contains 1,449 images with object instance labels. By removing the unlabelled regions near the image boundary,  a subset of 400 test images with size 608$\times$448 are collected for superpixel evaluation. As in conventions \cite{yang2020superpixel}, we directly apply the models of SSFCN \cite{yang2020superpixel}, LNSnet \cite{zhu2021learning} and our \textsc{ClustSeg} trained on BSDS500 to 400 NYUv2 images without any fine-tuning, to test the generalizability of the learning-based methods.

	\textbf{Training.} For superpixel query-initialization, we use a grid sampler to automatically sample a specified number of position-embedded pixel features as superpixel seeds. The network is trained jointly with the smooth L1 loss, and SLIC loss \cite{yang2020superpixel}. They are combined with coefficients of $10$ for smooth L1 and $1$ for SLIC losses.

    \textbf{Quantitative Results.} Fig.~\ref{fig:5} provides additional performance comparison of \textsc{ClustSeg} against both traditional (\ie, SLIC \cite{achanta2012slic}) and deep learning-based (\ie, SSFCN \cite{yang2020superpixel}, LNSnet \cite{zhu2021learning}) superpixel segmentation algorithms on NYUv2 \cite{silberman2012indoor} \texttt{test}. We can observe that \textsc{ClustSeg} consistently outperforms all the competitors in terms of ASA and CO. This also verifies stronger generalizability of \textsc{ClustSeg} over all the other learning-based competitors.

\begin{figure*}[!ht]
    \centering
    \includegraphics[width=\textwidth]{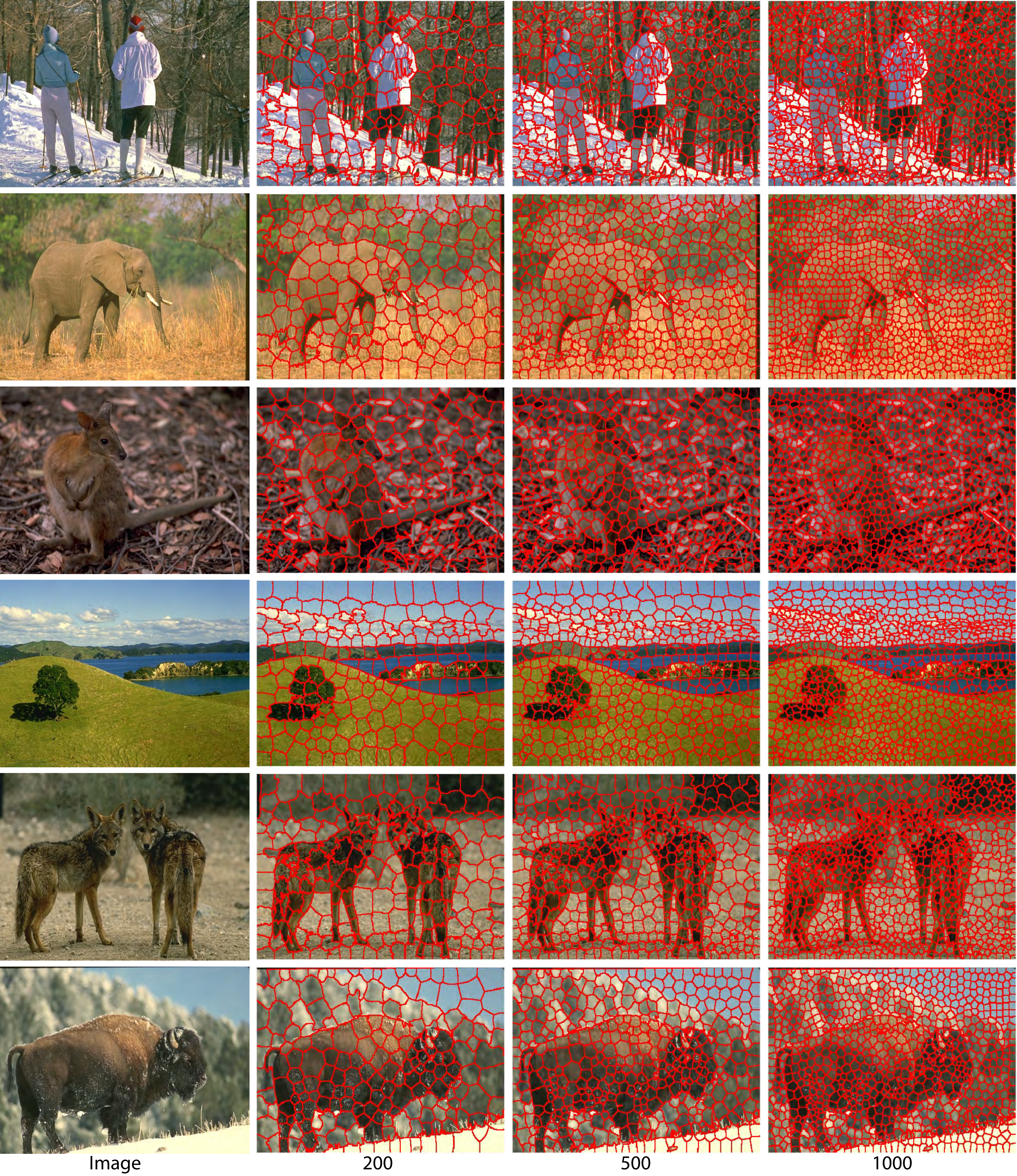}
    \caption{\textbf{Qualitative superpixel segmentation} results on BSDS500~\cite{arbelaez2011contour} \texttt{test}. For each test image, we show segmentation results with three different numbers of superpixels (\ie, 200, 500, and 1000). See \S\ref{sec:sup} for details.}
    \label{fig:sup}
\end{figure*}

\textbf{Qualitative Results.} Overall, \textsc{ClustSeg} can capture rich details in images and tends to create compact fine-grained results that closely align with object boundaries (see Fig.~\ref{fig:sup}). Across the different numbers of superpixels (\ie, 200 to 1000), \textsc{ClustSeg}  yields stable and impressive performance for various landscapes and objects.

\subsection{Failure Case Analysis}\label{sec:failure}
\begin{figure*}[!ht]
   \centering
   \includegraphics[width=0.9\textwidth]{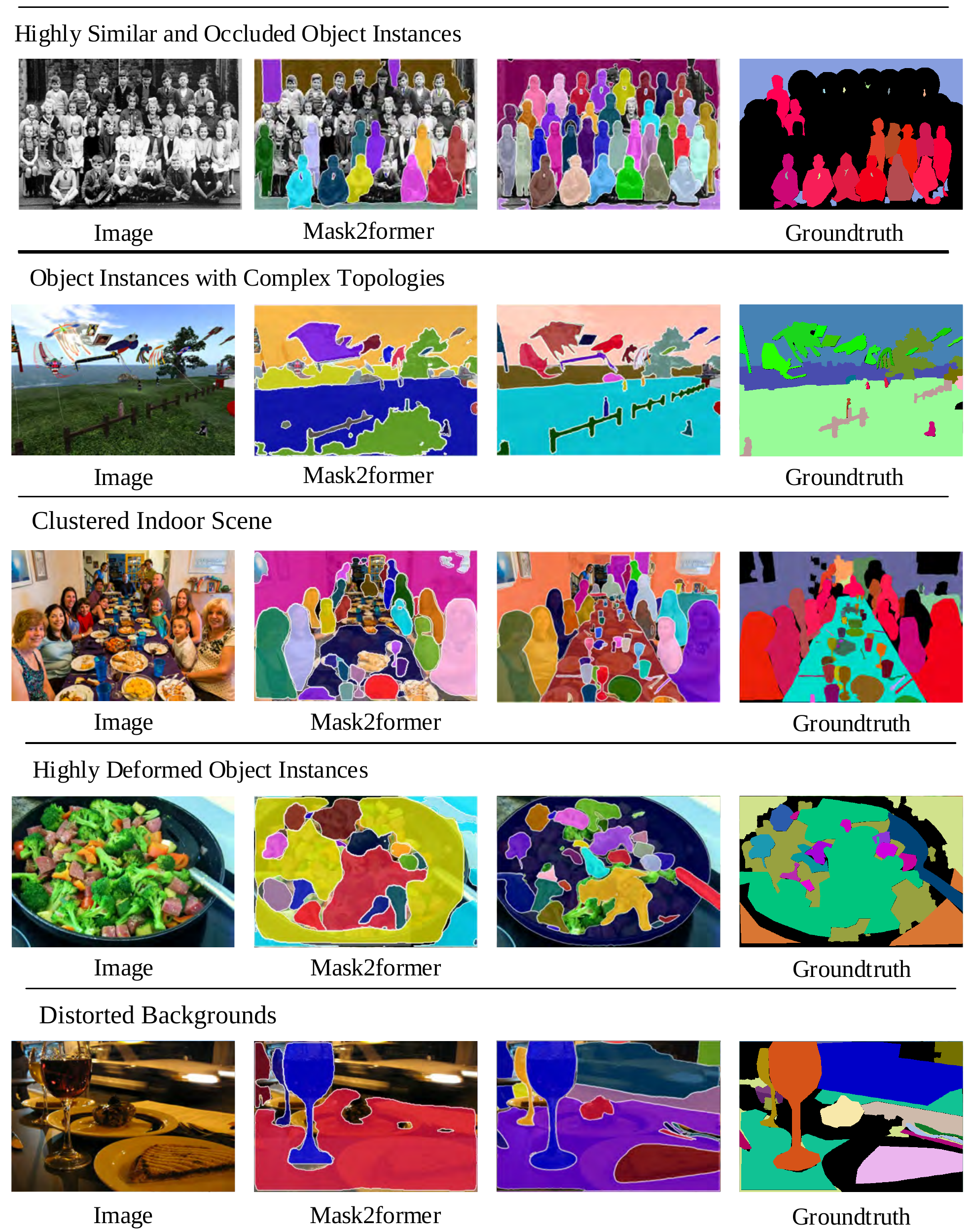}
       \put(-185,117){\scalebox{1.1}{\textsc{ClustSeg}}}
     \put(-185,5){\scalebox{1.1}{\textsc{ClustSeg}}}
     \put(-185,224){\scalebox{1.1}{\textsc{ClustSeg}}}
     \put(-185,339){\scalebox{1.1}{\textsc{ClustSeg}}}
     \put(-185,449){\scalebox{1.1}{\textsc{ClustSeg}}}
   \caption{\textbf{Failure Cases} on COCO panoptic~\cite{kirillov2019panoptic} \texttt{val}.  See \S\ref{sec:failure} for details.}
   \label{fig:failure}
\end{figure*}

As shown in Fig.~\ref{fig:failure}, we summarize the most representative failure cases and draw conclusions regarding their characteristic patterns that can lead to subpar results. Observedly, our algorithm struggles to separate objects from backgrounds in a number of incredibly complex scenarios (\ie, highly similar and occluded instances, objects with complex topologies, small objects, highly deformed objects, and distorted backgrounds). Developing more robust and powerful clustering algorithms may help alleviate these issues.

\section{More Ablative Studies} \label{sec:AS}
In this section, we provide more ablative studies regarding \textit{Dreamy-Start} query-initialization in Algorithm~\ref{alg:ini} and \textit{Recurrent Cross-Attention} for recursive clustering in Algorithm~\ref{alg:update}.
\subsection{Recurrent Cross-Attention}\label{sec:b1}
We perform further ablation studies on our non-recurrent cross-attention for the panoptic segmentation task. The results are summarized in the table below, where PQ (\%) is reported. As seen, simply stacking multiple non-recurrent cross-attention layers cannot achieve similar performance to our recurrent cross-attention with the same number of total iterations. Note that using multiple non-recurrent cross-attention layers even causes extra learnable parameters. EM is an iterative computational procedure for progressively estimating the local representatives of data samples in a given embedding space. When using multiple non-recurrent cross-attention layers, we essentially conduct one-step clustering on different embedding spaces, since the parameters are not shared among different cross-attention layers. This does not follow the nature of EM clustering, hence generating inferior results.

\begin{table}[h]
\vspace{-.5cm}
\centering
\caption{Ablative study of \textbf{recurrent cross-Attention $vs.$ non-recurrent cross-attention} over ResNet-50~\cite{he2016deep} on COCO Panoptic~\cite{kirillov2019panoptic}~\texttt{val} (see \S\ref{sec:b1} for details).}
\hspace{-1em}
						\setlength\tabcolsep{2pt}
						\renewcommand\arraystretch{1.02}
						\begin{tabular}{c|c|c||c}
							\thickhline
							\rowcolor{mygray}
							Iteration (T) & Recurrent cross-attention & Multiple non-recurrent	 & Additional learnable parameter \\
							\hline
                                \hline
							1 & 53.8 & 53.8 & - \\
                                2 & 54.1 & 53.8 & 1.3M \\
                                3 & 54.3 & 53.9 & 2.8M \\
                                4 & 54.3 & 53.9 & 4.3M \\
                                5 & 54.3 & 54.0 & 5.7M \\
							\hline
					\end{tabular}
					\label{B1}
     \vspace{-.5cm}
				\end{table}

\subsection{Query Initialization} \label{sec:b2}
We report the panoptic segmentation results with more iterations when learning queries as free parameters. As seen in Tab.~\ref{B2}, when learning initial queries as free parameters, even if using more iterations, performance degradation is still observed. Actually, the performance of iterative clustering algorithms heavily relies on the selection of initial seeds due to their stochastic nature~\cite{hamerly2002alternatives,celebi2013comparative,khan2004cluster}. This issue, called initial starting conditions, has long been a focus in the field of data clustering. It is commonly recognized that the effect of initial starting conditions cannot be alleviated by simply using more iterations. And this is why many different initialization methods are developed for more effective clustering~\cite{khan2004cluster}.

\begin{table}[ht]
\centering
\caption{Ablative study of \textbf{query initialization} over ResNet-50~\cite{he2016deep} on COCO Panoptic~\cite{kirillov2019panoptic}~\texttt{val} (see \S\ref{sec:b2} for details).}
\hspace{-1em}
						\setlength\tabcolsep{2pt}
						\renewcommand\arraystretch{1.02}
						\begin{tabular}{c|c|ccc|cc}
							\thickhline
							\rowcolor{mygray}
							Method & Iteration (T) & \text{PQ}  &$\text{PQ}^\text{Th}$ &$\text{PQ}^\text{St}$ &$\text{AP}^\text{Th}_\text{pan}$ &$\text{mIoU}_\text{pan}$ \\
							\hline
                                \hline
							\textbf{Dreamy-Start} & 3 & 54.3 & 60.4 & 45.8 & 42.2 & 63.8 \\
                                Free parameters & 3 & 53.5 & 59.6 & 45.1 & 41.0 & 60.5 \\
                                Free parameters & 3 & 53.7 & 59.9 & 45.3 & 54.2 & 61.1 \\
                                Free parameters & 3 & 53.8 & 60.1 & 45.4 & 41.6 & 61.4 \\
							\hline
					\end{tabular}
					\label{B2}
				\end{table}

\subsection{Deep Supervision}\label{sec:b3}
\begin{figure*}[!ht]
    \centering
    \includegraphics[width=\textwidth]{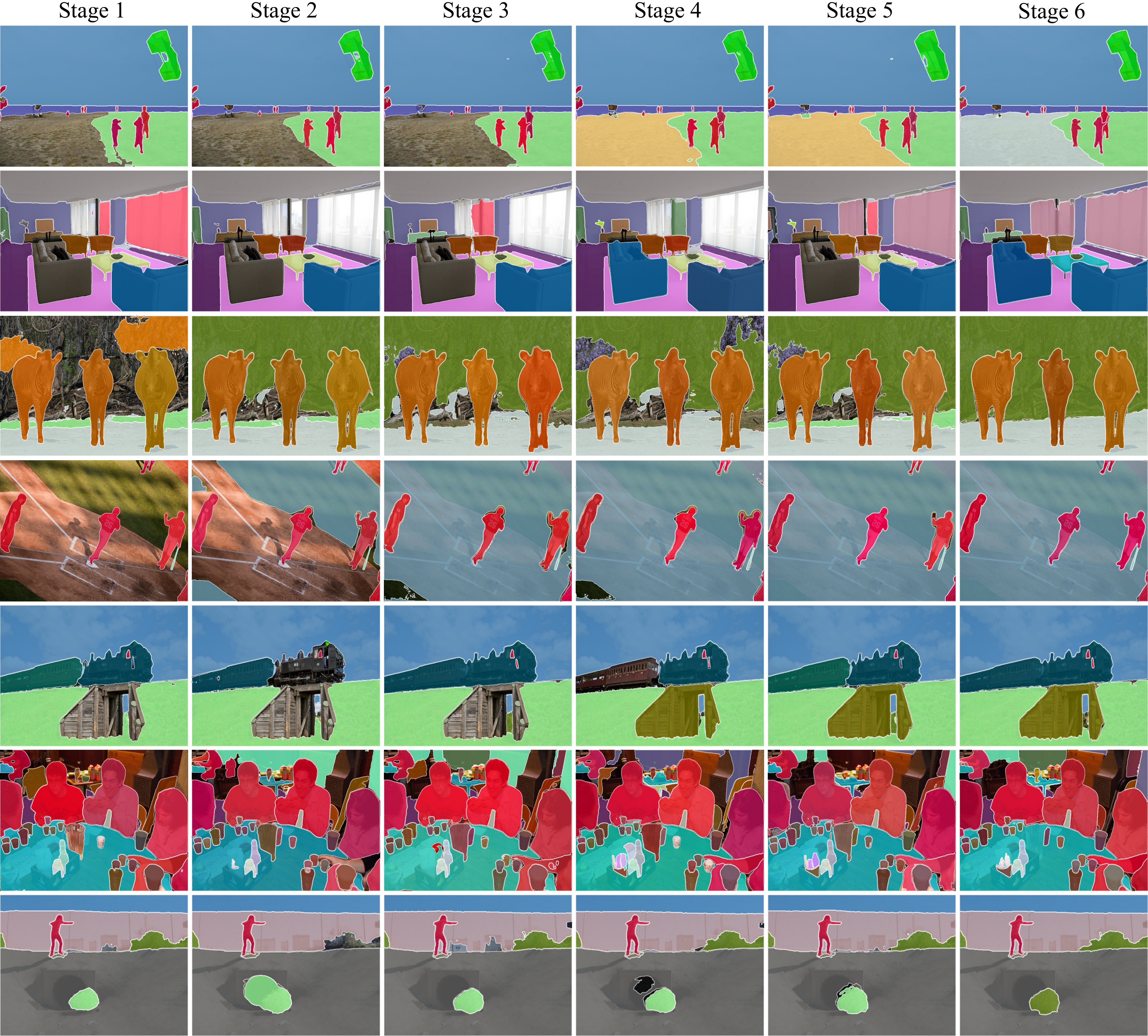}
    \caption{\textbf{Visualization of panoptic segmentation in different stages} results on COCO Panoptic~\cite{kirillov2019panoptic} \texttt{val} with \textsc{ClustSeg} with Swin-B~\cite{liu2021swin} backbone.  See \S\ref{sec:b3} for details.}
    \label{fig:b3}
\end{figure*}

We adopt deep supervision to train every E-step of each recurrent cross-attention. A similar strategy is widely employed in previous segmentation models and other Transformer counterparts, \eg, Mask2Former~\cite{cheng2021masked}, kMaX-Deeplab~\cite{yu2022k}. We ablate the effect of such a deep supervision strategy for panoptic segmentation in Tab.~\ref{table:b3a}. Moreover, we also show the accuracy of segmentation predictions from different iterations of the last recurrent cross-attention layer in Tab.~\ref{table:b3b}. We additionally provide visualization of segmentation results in different stages in Fig.~\ref{fig:b3}.

\begin{table*}[ht]
\vspace{-7px}
					\caption{Ablative studies of \textbf{deep supervision} over ResNet-50~\cite{he2016deep} on COCO Panoptic~\cite{li2022panoptic} \texttt{val} (see \S\ref{sec:b3}).}
					\begin{subtable}{0.56\linewidth}
						\captionsetup{width=.95\linewidth}
						\resizebox{\textwidth}{!}{
							\setlength\tabcolsep{4pt}
							\renewcommand\arraystretch{1.1}
							\begin{tabular}{c|ccc|cc}
								\thickhline
								\rowcolor{mygray}
								Variant & \text{PQ}  &$\text{PQ}^\text{Th}$ &$\text{PQ}^\text{St}$ &$\text{AP}^\text{Th}_\text{pan}$ &$\text{mIoU}_\text{pan}$ \\
								\hline\hline
								\arrayrulecolor{gray}\hdashline\arrayrulecolor{black}	
								\makecell{Only final E-step of \\each recurrent cross-attention	}& 53.0 & 59.6 & 43.7 & 41.7 & 61.2 \\ 	
								Deep supervision & 54.3 & 60.4 & 45.8 & 42.2 & 63.8 \\
								\hline
						\end{tabular}}
						\setlength{\abovecaptionskip}{0.3cm}
						\setlength{\belowcaptionskip}{-0.1cm}
						\caption{\small{Supervision variants}}
						\label{table:b3a}
					\end{subtable}
                    \vspace{-6px}
					\hspace{-0.5em}
					\begin{subtable}{0.44\linewidth}
						\resizebox{\textwidth}{!}{
							\setlength\tabcolsep{4pt}
							\renewcommand\arraystretch{1.1}
							\begin{tabular}{c|ccc|cc}
								\thickhline
								\rowcolor{mygray}
								Iteration (T) & \text{PQ}  &$\text{PQ}^\text{Th}$ &$\text{PQ}^\text{St}$ &$\text{AP}^\text{Th}_\text{pan}$ &$\text{mIoU}_\text{pan}$\\
								\hline\hline
								1 & 53.8 & 59.7 & 45.6 & 41.6 & 63.1\\
								2 & 54.0 & 60.1 & 45.6 & 41.9 & 63.4\\
                                3 & 54.3 & 60.4 & 45.8 & 42.2 & 63.8\\
								\hline
							\end{tabular}
						}
						\setlength{\abovecaptionskip}{0.3cm}
						\setlength{\belowcaptionskip}{-0.1cm}
						\caption{\small{Iterations of the last recurrent cross-attention layer}}
						\label{table:b3b}
					\end{subtable}
					\vspace{-20pt}
				\end{table*}

\section{Pseudo Code} \label{sec:PC}
In this section, we provide pseudo-code of \textit{Dreamy-Start} query-initialization in Algorithm~\ref{alg:ini} and \textit{Recurrent Cross-Attention} for recursive clustering in Algorithm~\ref{alg:update}.

	\begin{algorithm}[t]
	\caption{Pseudo-code of \textit{Dreamy-Start} for query initialization in a PyTorch-like style.}
	\label{alg:ini}
	\definecolor{codeblue}{rgb}{0.25,0.5,0.5}
	\lstset{
		backgroundcolor=\color{white},
		basicstyle=\fontsize{7.8pt}{7.8pt}\ttfamily\selectfont,
		columns=fullflexible,
		breaklines=true,
		captionpos=b,
		escapeinside={(:}{:)},
		commentstyle=\fontsize{7.8pt}{7.8pt}\color{codeblue},
		keywordstyle=\fontsize{7.8pt}{7.8pt},
	}
	\begin{lstlisting}[language=python]
"""
feats: output feature of backbone, shape: (channels, height, width)
memory: a set of queues storing class-aware pixel embeddings, each has a shape of (num_feats, channels)
num_sp: number of superpixels
FFN: feedforward network, PE: position embedding
"""

# scene-agnostic center initialization (Eq.7)
(:\color{codedefine}{\textbf{def}}:) (:\color{codefunc}{\textbf{scene\_agnostic\_initialization}}:)(memory):

    mem_feats = (:\color{codefunc}{\textbf{Avg\_Pool}}:)(memory)

    semantic_centers = (:\color{codefunc}{\textbf{FFN}}:)(mem_feats)

   (:\color{codedefine}{\textbf{return}}:) semantic_centers

# scene-adaptive center initialization (Eq.8)
(:\color{codedefine}{\textbf{def}}:) (:\color{codefunc}{\textbf{scene\_adaptive\_initialization}}:)(feats):

    feats = (:\color{codefunc}{\textbf{PE}}:)(feats)

    instance_centers = (:\color{codefunc}{\textbf{FFN}}:)(feats)

    (:\color{codedefine}{\textbf{return}}:) instance_centers

# superpixel center initialization (Eq.9)
(:\color{codedefine}{\textbf{def}}:) (:\color{codefunc}{\textbf{superpixel\_initialization}}:)(feats):

    _, H, W = feats.(:\color{codepro}{\textbf{shape}}:)

    feats = (:\color{codefunc}{\textbf{PE}}:)(feats)

    # Grid sampler of num_sp superpixels
    f = (:\color{codefunc}{\textbf{torch.sqrt}}:)(num_sp/H/W)
    x = (:\color{codefunc}{\textbf{torch.linspace}}:)(0, W, (:\color{codefunc}{\textbf{torch.int}}:)(W*f))
    y = (:\color{codefunc}{\textbf{torch.linspace}}:)(0, H, (:\color{codefunc}{\textbf{torch.int}}:)(H*f))
    meshx, meshy = (:\color{codefunc}{\textbf{torch.meshgrid}}:)((x, y))
    grid = (:\color{codefunc}{\textbf{torch.stack}}:)((meshy, meshx), 2).(:\color{codepro}{\textbf{unsqueeze}}:)(0)
    feats = (:\color{codefunc}{\textbf{grid\_sample}}:)(feats, grid).(:\color{codepro}{\textbf{view}}:)(-1, channels)

    superpixel_centers = (:\color{codefunc}{\textbf{FFN}}:)(feats)

    (:\color{codedefine}{\textbf{return}}:) superpixel_centers
        \end{lstlisting}
\end{algorithm}

	\begin{algorithm}[t]
	\caption{Pseudo-code of \textit{Recurrent Cross-attention} for \textit{Recursive Clustering} in a PyTorch-like style.}
	\label{alg:update}
	\definecolor{codeblue}{rgb}{0.25,0.5,0.5}
	\lstset{
		backgroundcolor=\color{white},
		basicstyle=\fontsize{7.8pt}{7.8pt}\ttfamily\selectfont,
		columns=fullflexible,
		breaklines=true,
		captionpos=b,
		escapeinside={(:}{:)},
		commentstyle=\fontsize{7.8pt}{7.8pt}\color{codeblue},
		keywordstyle=\fontsize{7.8pt}{7.8pt},
	}
	\begin{lstlisting}[language=python]
"""
feats: output feature of backbone, shape: (batch_size, channels, height, width)
C: cluster centers, shape: (batch_size, num_clusters, dimention)
T: iteration number for recursive clustering
"""

# One-step cross attention in Eq.10
(:\color{codedefine}{\textbf{def}}:) (:\color{codefunc}{\textbf{recurrent\_cross\_attention\_layer}}:)(Q, K, V):

    # E-step
    output = (:\color{codefunc}{\textbf{torch.matmul}}:)(Q, K.(:\color{codepro}{\textbf{transpose}}:)(-2, -1))
    M = (:\color{codefunc}{\textbf{torch.nn.functional.softmax}}:)(output, (:\color{codedim}{\textbf{dim}}:)=-2)

    # M-step
    C = (:\color{codefunc}{\textbf{torch.matmul}}:)(M, V)

    (:\color{codedefine}{\textbf{return}}:) C

# Recurrent cross-attention in Eq.11
(:\color{codedefine}{\textbf{def}}:) (:\color{codefunc}{\textbf{RCross\_Attention}}:)(feats, C, T):

    Q = (:\color{codefunc}{\textbf{nn.Linear}}:)(C)
    K = (:\color{codefunc}{\textbf{nn.Linear}}:)(feats)
    V = (:\color{codefunc}{\textbf{nn.Linear}}:)(feats)
    C = (:\color{codefunc}{\textbf{recurrent\_cross\_attention\_layer}}:)(Q, K, V)

    (:\color{codedefine}{\textbf{for}}:) _ (:\color{codedefine}{\textbf{in}}:) (:\color{codedefine}{range}:)(T-1):
        Q = (:\color{codefunc}{\textbf{nn.Linear}}:)(C)
        C = (:\color{codefunc}{\textbf{recurrent\_cross\_attention\_layer}}:)(Q, K, V)

    (:\color{codedefine}{\textbf{return}}:) C

	\end{lstlisting}
\end{algorithm}

\section{Discussion}
\label{sec:dis}

\textbf{Asset License and Consent.}
We apply five closed-set image segmentation datasets, \ie, MS COCO~\cite{lin2014microsoft}, MS COCO Panoptic~\cite{kirillov2019panoptic}, ADE20K~\cite{zhou2017scene}, BSDS500~\cite{arbelaez2011contour} and NYUv2~\cite{silberman2012indoor}
They are all publicly and freely available for academic purposes. We implement all models with MMDetection~\cite{contributors2019openmmlab}, MMSegmentation~\cite{contributors2020mmsegmentation} and Deeplab2~\cite{chen2017deeplab,max_deeplab_2021,yu2022k} codebases. MS COCO (\href{https://cocodataset.org/}{https://cocodataset.org/}) is released under a \href{https://creativecommons.org/licenses/by/4.0/legalcode}{CC BY 4.0}; MS COCO Panoptic (\href{https://github.com/cocodataset/panopticapi}{https://github.com/cocodataset/panopticapi}) is released under a \href{https://creativecommons.org/licenses/by/4.0/legalcode}{CC BY 4.0}; ADE20K (\href{https://groups.csail.mit.edu/vision/datasets/ADE20K/}{https://groups.csail.mit.edu/vision/datasets/ADE20K/}) is released under a \href{https://opensource.org/licenses/BSD-3-Clause}{CC BSD-3}; All assets mentioned above release annotations obtained from human experts with agreements. MMDetection (\href{https://github.com/open-mmlab/mmdetection}{https://github.com/open-mmlab/mmdetection}),  MMSegmentation (\href{https://github.com/open-mmlab/mmsegmentation}{https://github.com/open-mmlab/mmsegmentation}) and Deeplab2 codebases (\href{https://github.com/google-research/deeplab2}{https://github.com/google-research/deeplab2}) are released under \href{https://www.apache.org/licenses/LICENSE-2.0}{Apache-2.0}.

\textbf{Limitation Analysis.}  One limitation of our algorithm arises from the extra clustering loops in each training iteration, as they may reduce the computation efficiency in terms of time complexity. However, in practice, we observe that three recursive clusterings are sufficient for global model convergence, incurring only a minor computational overhead, \ie, ~5.19\% reduction in terms of training speed. We will dedicate ourselves to the development of potent algorithms that are more efficient and effective.

\textbf{Broader Impact.} This work develops a universal and transparent segmentation framework, which unifies different image segmentation tasks from a clustering perspective. We devise a novel cluster center initialization scheme as well as a neural solver for iterative clustering, hence fully exploiting the fundamental principles of recursive clustering for pixel grouping.  Our algorithm has demonstrated its effectiveness over a variety of famous models in four core segmentation tasks (\ie, panoptic, instance,  semantic, and superpixel segmentation). On the positive side, our approach has the potential to benefit a wide variety of applications in the real world, such as autonomous vehicles, robot navigation, and medical imaging. On the other side, erroneous predictions in real-world applications (\ie, medical imaging analysis and any tasks involving autonomous vehicles) give rise to concerns about the safety of human beings. In order to avoid this potentially negative effect on society and the community, we suggest proposing a highly stringent security protocol in the event that our approach fails to function properly in real-world applications.

\end{document}